\documentclass{article}

\usepackage{arxiv}
\usepackage{times}
\usepackage{epsfig}
\usepackage{graphicx}
\usepackage{amsmath}
\usepackage{amssymb}

\usepackage{caption}
\usepackage{subcaption}
\usepackage{multirow}
\usepackage{multicol}
\usepackage{booktabs}
\usepackage{amsfonts} 
\usepackage{nicefrac}
\usepackage{enumitem}
\newtheorem{theorem}{Theorem}

\usepackage{amsmath,amsfonts,bm}

\def\eqref#1{equation~\ref{#1}}

\def\1{\bm{1}}

\def\rx{{\textnormal{x}}}

\def\rz{{\textnormal{z}}}

\def\vmu{{\bm{\mu}}}

\def\vc{{\bm{c}}}

\def\vn{{\bm{n}}}

\def\vp{{\bm{p}}}

\def\vz{{\bm{z}}}

\def\mF{{\bm{F}}}

\def\mI{{\bm{I}}}

\def\mM{{\bm{M}}}

\def\mP{{\bm{P}}}

\def\mR{{\bm{R}}}

\def\mT{{\bm{T}}}
\def\mU{{\bm{U}}}
\def\mV{{\bm{V}}}
\def\mW{{\bm{W}}}

\def\mZ{{\bm{Z}}}

\def\mSigma{{\bm{\Sigma}}}

\DeclareMathAlphabet{\mathsfit}{\encodingdefault}{\sfdefault}{m}{sl}
\SetMathAlphabet{\mathsfit}{bold}{\encodingdefault}{\sfdefault}{bx}{n}

\newcommand{\E}{\mathbb{E}}

\newcommand{\R}{\mathbb{R}}

\newcommand{\KL}{D_{\mathrm{KL}}}

\DeclareMathOperator*{\argmin}{arg\,min}

\def\M{\mathcal{M}}

\usepackage[pagebackref=true,breaklinks=true,colorlinks,bookmarks=false]{hyperref}

\def\cvprPaperID{7742} 
\def\confYear{CVPR 2021}

\begin{document}

\title{Unsupervised Geometric Disentanglement for Surfaces via CFAN-VAE}

\author{
  N. Joseph Tatro  \\
  Department of Mathematical Sciences\\
  Rensselaer Polytechnic Institute\\
  Troy, NY 12180 \\
  \texttt{tatron@rpi.edu} \\
  \And
 Stefan C. Schonsheck \\
  Department of Mathematical Sciences\\
  Rensselaer Polytechnic Institute\\
  Troy, NY 12180 \\
  \texttt{schons@rpi.edu} \\
 \And
 Rongjie Lai \\
  Department of Mathematical Sciences\\
  Rensselaer Polytechnic Institute\\
  Troy, NY 12180 \\
  \texttt{lair@rpi.edu} \\
}

\maketitle

\begin{abstract}
   Geometric disentanglement, the separation of latent codes for intrinsic (i.e. identity) and extrinsic (i.e. pose) geometry, is a prominent task for generative models of non-Euclidean data such as 3D deformable models. It provides greater interpretability of the latent space, and leads to more control in generation. This work introduces a mesh feature, the conformal factor and normal feature (CFAN), for use in mesh convolutional autoencoders. We further propose CFAN-VAE, a novel architecture that disentangles identity and pose using the CFAN feature. Requiring no label information on the identity or pose during training, CFAN-VAE achieves geometric disentanglement in an unsupervised way. Our comprehensive experiments, including reconstruction, interpolation, generation, and identity/pose transfer, demonstrate CFAN-VAE achieves state-of-the-art performance on unsupervised geometric disentanglement. We also successfully detect a level of geometric disentanglement in mesh convolutional autoencoders that encode xyz-coordinates directly by registering its latent space to that of CFAN-VAE.
\end{abstract}


\section{Introduction}

Of recent interest in the deep learning community, generative models have proved to be powerful tools for many tasks including synthetic data generation and style transfer \cite{goodfellow2014generative}. Geometric deep learning is a new field interested in extending such success of deep learning to non-Euclidean structured data \cite{Bronstein_2017}. The development of this field is timely given the recent proliferation of point cloud and mesh structured data obtained from sources such as laserscanners \cite{Geiger2013IJRR} and CAD software \cite{chang2015shapenet}. 

Particularly, mesh based convolutional autoencoders (MeshVAEs) are now a popular tool for generating surfaces \cite{cheng2019meshgan, Litany_2018, Ma2019Dressing3H, Ranjan_2018}. These models process a surface via geometric convolutions that respect its intrinsic geometry. 
With these VAEs achieving state-of-the-art performance on tasks such as reconstruction, more attention is being given towards tasks such as latent space interpretability. Geometric disentanglement, where the latent variables controlling intrinsic (properties independent of surface embedding) and extrinsic (properties dependent on surface embedding) geometry are separated \cite{aumentadoarmstrong2019geometric}, is an important open problem related to such interpretability. Applications include graphics, where a typical example is a disentangled latent space separating identity and pose in the case of human body generation \cite{jiang2019learning, Tan_2018}.

Typically, MeshVAEs encode a surface using an input feature that couples intrinsic and extrinsic geometry. In the common case where 3D coordinates are directly encoded, we refer to such networks as xyz-VAEs. Given that it is learned from encoding a feature that entangles intrinsic and extrinsic geometry, the xyz-VAE latent space may be geometrically entangled. In this paper, we are interested in creating an architecture that explicitly leads to geometric disentanglement in an unsupervised setting. Namely, we do not require labels for mesh identity and pose. This allows the architecture to be applied more broadly as it is either quite time consuming or sometimes even unrealistic to have meaningful identity and pose labels in many mesh datasets.

\begin{figure*}[tb]
\centering
  \begin{subfigure}[b]{0.29\textwidth}
    \includegraphics[width=\textwidth]{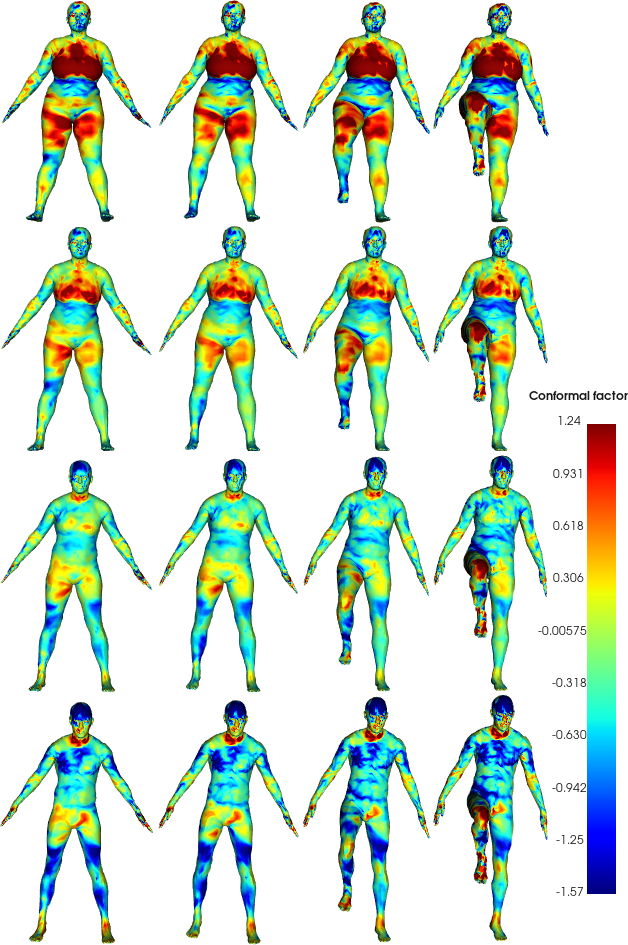}
  \end{subfigure}
\hfill
  \begin{subfigure}[b]{0.25\textwidth}
    \includegraphics[width=\textwidth]{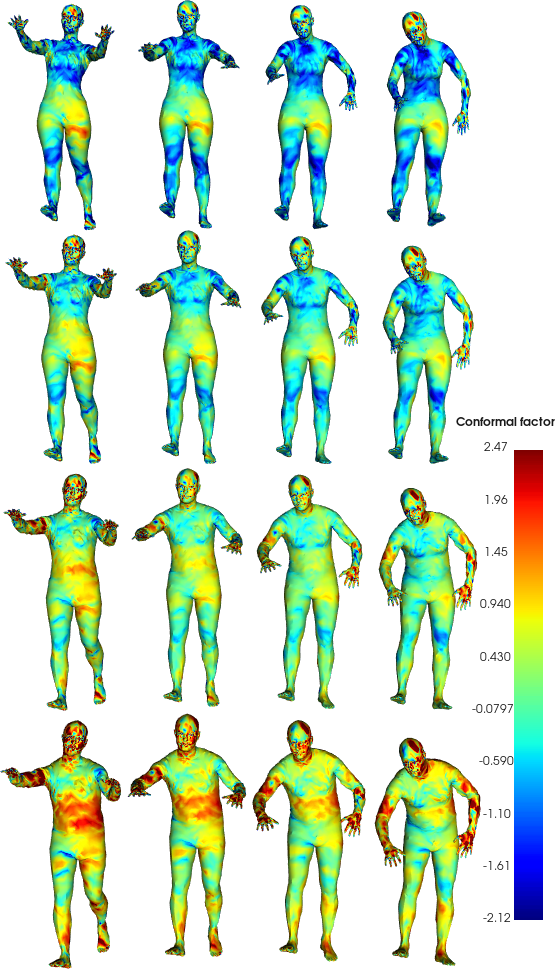}
    \end{subfigure}
\hfill
\begin{subfigure}[b]{0.40\textwidth}
    \includegraphics[width=\textwidth]{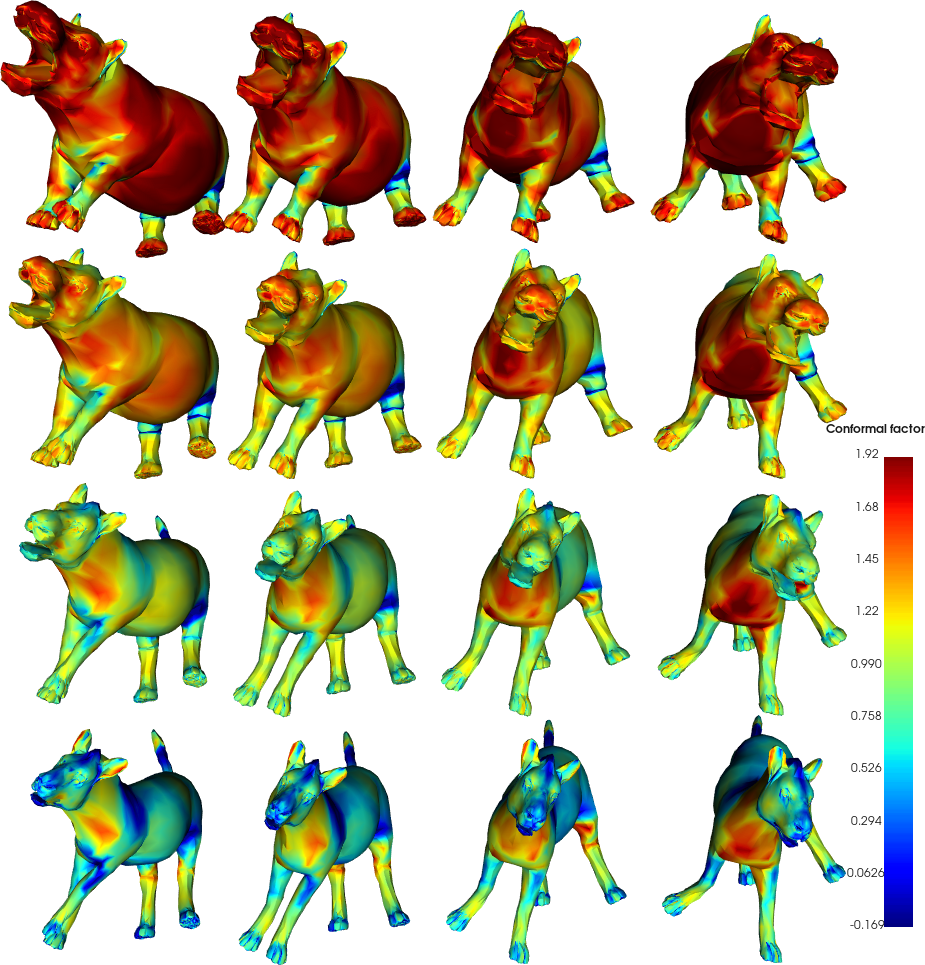}
\end{subfigure}
\caption{Geometric disentangled interpolations between two meshes from DFAUST, SURREAL, and SMAL datasets. Meshes are generated by CFAN-VAEs. The horizontal/vertical axis display linear interpolation in the normal/conformal latent codes, $\vz_n$/$\vz_c$. Color denotes the conformal factor feature of each reconstructed mesh. 
}
\label{fig:interp}
\end{figure*}

\subsection{Contributions}
To summarize our contributions, in this work, we:
\begin{enumerate}[nolistsep]
\setlength\itemsep{0em}
    \item  introduce a feature, the conformal factor and normal feature (CFAN), that decouples intrinsic and extrinsic geometry for use in mesh convolutional autoencoders. 
    \item propose a novel architecture, CFAN-VAE, for unsupervised geometric disentanglement. For a given mesh, we compute the CFAN feature, and encode its components separately into latent vectors representing intrinsic and extrinsic geometry. We then jointly decode these vectors to the 3D coordinates of the mesh. 
    \item investigate geometric disentanglement in generic xyz-VAEs by registering this latent space to the disentangled latent space of CFAN-VAE via solving the \textit{Orthogonal Procrustes (OP) problem}. We recover a disentangled latent space for xyz-VAEs without any additional conditions or labels. 
    \item perform comprehensive numerical experiments using three datasets, achieving state-of-the-art performance in unsupervised geometric disentanglement.
\end{enumerate}

The rest of this work is organized as follows; we first discuss related work. Then we introduce our mesh signal, the CFAN feature, and novel architecture, CFAN-VAE. After this, we describe the process of latent space registration for comparing the different latent spaces. Finally, we detail our numerical experiments including reconstruction, interpolation, generation, and latent space analysis.

\subsection{Related Work}

\paragraph{Geometric Convolutional Generative Models} Until fairly recently, the standard technique for generating 3D shapes was through the use of volumetric convolutional neural networks (CNNs) which act on 3D voxels \cite{wu2016learning, Zhirong_Wu_2015}. However, the use of voxels yields coarse representations. Other work has investigated generating surfaces through utilizing point cloud representations. In \cite{achlioptas2017learning}, point clouds of fixed size are generated utilizing the PointNet architecture \cite{Charles_2017}. A drawback of point clouds is a lack in connectivity between points, thus the structure is not necessarily smooth. 

Using non-Euclidean convolutions has shown great potential to overcome limitations due to 3D voxel or point cloud representation. These non-Euclidean convolutions include spectral methods \cite{bruna2013spectral, defferrard2016convolutional, kipf2016semisupervised}, patch based methods \cite{boscaini2016learning, jin2019nptc, Masci_2015, Monti_2017, schonsheck2018parallel}, and hard-attention methods \cite{bouritsas2019neural}. 

There has been recent research into the construction of generative models using these non-Euclidean convolutions. \cite{Litany_2018} first proposed using VAEs to generate 3D meshes. Their proposed network can be used for mesh completion in human bodies, essentially the inpainting problem extended to surfaces. Recently, \cite{Ranjan_2018} used a MeshVAE to generate realistic human faces. There has also been work in developing generative adversarial networks (MeshGANs); \cite{cheng2019meshgan} uses a MeshGAN to generate meshes of human faces at high resolution and \cite{Ma2019Dressing3H} generates clothes on 3D human bodies.

\paragraph{Geometric Disentanglement} Geometric disentanglement is an important feature leading to a more interpretable latent space and thus more control in generation. 
For instance, \cite{jiang2019learning, jiang2019disentangled, Tan_2018} learn a latent space separating identity and pose, advocating for the use of the \textit{as-consistent-as-possible deformation representation} (ACAPDR) introduced in \cite{Gao_2019} instead of 3D coordinates. These works require a reference pose for all identities during training. Recently, several works proposed networks for geometric disentanglement using non-Euclidean convolution \cite{cosmo2020limp, levinson2019latent, zhou2020unsupervised}. The disentanglement in these works either includes supervision or loss functions that require identification of isometric pairs in training data. In practice, the latter results in needing training data with identity categorization, and thus supervised identity, to scale on large datasets. Training datasets lacking isometries may be desireable in applications such as differential privacy \cite{abadi2016deep}.  

There is limited work to achieve unsupervised geometric disentanglement that \textit{completely} eliminates reliance on both identity and pose labeling. To our knowledge, the only previous work concerning unsupervised disentanglement for 3D data is \cite{aumentadoarmstrong2019geometric}, in which the authors introduce GDVAE, a VAE built on a PointNet architecture \cite{Charles_2017}. The PointNet embedding of each point cloud is encoded into a split latent space, representing intrinsic and extrinsic features. The latent variables are decoded to recover the PointNet features from which the point cloud can be reconstructed. To promote disentanglement, a decoder on the intrinsic latent space is trained to predict the Laplace-Beltrami eigenvalues. 


\begin{figure*}[tb]
    \centering
    \includegraphics[width=0.75\textwidth]{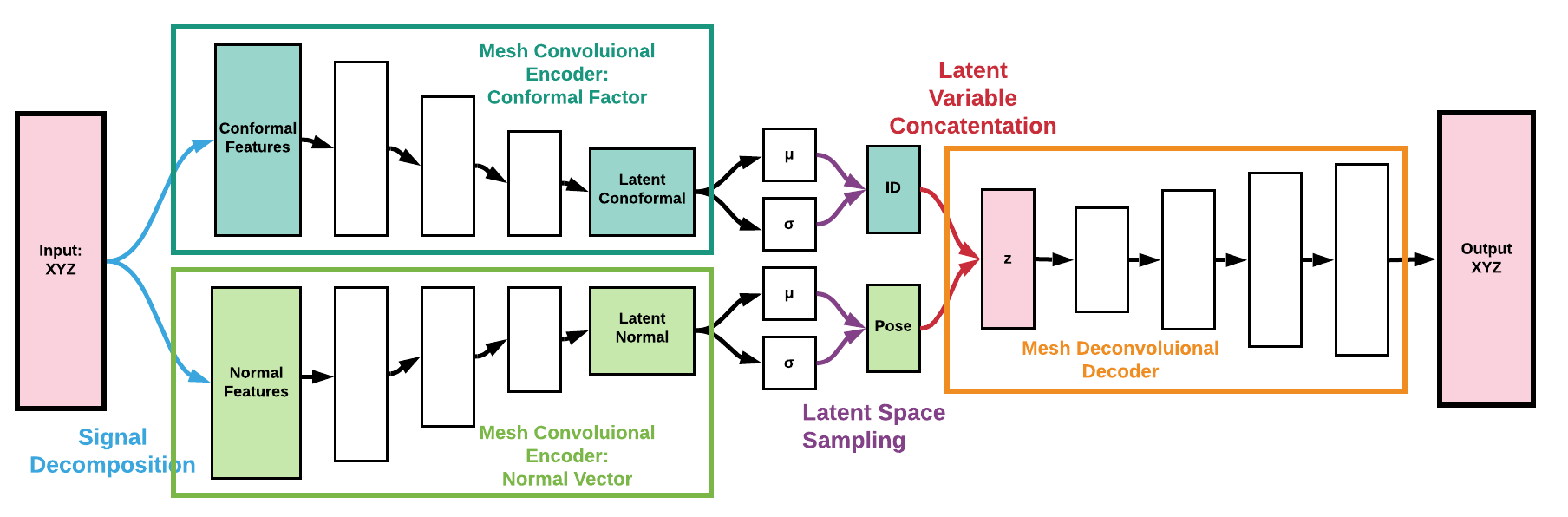}
    \caption{Diagram of the CFAN-VAE architecture. First, 3D coordinates are transformed into the CFAN feature. This feature is split and encoded to create conformal and normal latent variables, which are jointly decoded to reconstruct the 3D coordinates. The conformal and normal latent variables control identity and pose respectively, leading to disentanglement.}
    \label{fig:conformalvae_architecture}
\end{figure*}

\section{CFAN-VAE}

Our novel method for geometric disentanglement is motivated by the Fundamental Theorem of Surfaces \cite{do2016differential}. We describe surfaces using conformal factors and surface normal vectors. 
This CFAN feature is easily leveraged by CFAN-VAE, the network architecture we introduce, for geometric disentanglement. 
In addition, we use parallel transport convolution (PTC) introduced in~\cite{schonsheck2018parallel}, as it defines convolution through parallel transport which respects the intrinsic geometry of the surface. 

\subsection{CFAN Feature}
The Fundamental Theorem of Surfaces states that a surface can be uniquely reconstructed up to rigid motion given its metric tensor and surface normals, if they satisfy structural conditions known as the Gauss-Codazzi equations~\cite{do2016differential}. In essence, the metric tensor encapsulates the intrinsic geometry of a surface, while the normals encapsulates the extrinsic. Then we can use metric tensors and normal vector fields to characterize surfaces. 

In this work, we consider genus zero surfaces. It is well known that all genus zero surfaces are conformally equivalent~\cite{jost2008riemannian}. Namely, given two genus-zero surfaces, $(\mathcal{M}_1, g_1)$ and $(\mathcal{M}_2, g_2)$, there exists a diffeomorphism, 
\begin{align}
    \phi:(\mathcal{M}_1, g_1) \rightarrow (\mathcal{M}_2, g_2) \text{ s.t. }  \phi^*(g_2) = \exp(2 \lambda) g_1.
\end{align}
Here the function, $\lambda$, is known as the conformal factor and defines a conformal deformation from $\mathcal{M}_1$ to $\mathcal{M}_2$. 
Then any pair of conformally equivalent surfaces can be deformed into one another (up to isomorphism) by choosing the correct conformal factor. However, in order to properly reconstruct a surface embedded in 3D, we must also fix the isomorphism. This embedding is defined up to translation by the normal field. These two components define the CFAN feature, with the conformal factor and normal field representing intrinsic and extrinsic geometry respectively. 

\paragraph{Discretized CFAN}
Given a triangle mesh $(\mP, \mT)$, where $\mP \in \mathbb{R}^{n \times 3}$ is the set of vertices and $\mT \in \mathbb{R}^{k \times 3}$ is the corresponding set of faces, we define the discretized CFAN feature. Each face $\tau$ in $\mT$ has a corresponding exterior face normal $\vn_\tau$. We compute the weighted average of these face normals around the first ring structure of each vertex to define a pointwise normal.    
Since the local area on the surface is given by $\sqrt{\det g}$, it follows that the logarithm of local area is proportional to the conformal factor. 
Then we define the CFAN feature, $(c_i, \vn_i)$ as: 
\begin{align}
\label{eqn:CFAN}
    c_i = \log  \sum_{\substack{\tau \in \mT;\\ i \in \tau}} \frac{\text{Area}(\tau)}{3}, \qquad
    \vn_i = \frac{\sum_{\substack{\tau \in \mT;\\ i \in \tau}} \text{Area}(\tau) \vn_\tau}{||\sum_{\substack{\tau \in \mT;\\ i \in \tau}} \text{Area}(\tau) ||} 
\end{align}
Compared to popular features, such as ACAPDR or SHOT descriptors \cite{tombari2010unique}, CFAN is easier to compute and more compact. In practice, we perform a pointwise normalization of these features. Given sensitivity of normals to noise, Table \ref{tab:noise_ablation} in the appendix shows we are able to train CFAN-VAEs robust to noise on the vertices.

\subsection{Network Architecture}
Based on the CFAN feature, we propose a simple architecture, CFAN-VAE, as shown in Figure \ref{fig:conformalvae_architecture} to achieve unsupervised geometric disentanglement. 
The intuition is to encode the conformal factor and the normal features separately. The 3D coordinates $\vp$ of an input mesh under a fixed triangulation are first converted to the CFAN feature by transformations $\phi_c$ and $\phi_n$ provided by \eqref{eqn:CFAN}.
\begin{align}
    \vc = \phi_c (\vp), \qquad \vn = \phi_n (\vp).
\end{align}
Then, the feature components are separately encoded by $E_c$ and $E_n$ to create two different latent variables, the conformal latent variable $\vz_{c}$ and the normal latent variable $\vz_{n}$ in the disentangled latent space $\mZ_{c,n} = \mZ_{c}\times\mZ_{n}$,
\begin{align}
    \vz_c = E_c(\vc), \qquad \vz_n = E_n(\vn), \qquad \hat{\vp} = D(\vz_c, \vz_n). 
\end{align}
As a result, $\vz_{c}$ corresponds to intrinsic geometry, controlling surface identity, and $\vz_{n}$ corresponds to extrinsic geometry, controlling surface pose. After that, the CFAN latent variable $\vz_{c, n} = [\vz_{c},\vz_{n}]\in\mZ_{c,n}$ is decoded by $D$ to obtain the reconstructed 3D coordinates $\hat{\vp}$,

Our model is built on geometric convolutions given by PTC layers \cite{schonsheck2018parallel}. Details on PTC are included in Appendix \ref{subsubsec:PTC}. To follow convention for convolutional VAEs, the mesh signal is decimated during encoding and refined during decoding by a factor of four. 
Sampling is precomputed by the geodesic Farthest Point Sampling (FPS) method \cite{moenning2003fast}.
After the final mesh convolution in the encoder, dense layers map the signal to the variational statistics, $\mu := [\mu_c, \mu_n]$ and $\sigma := [\sigma_c, \sigma_n]$. A dense layer also maps $\vz_{c, n}$ to the first mesh signal in the decoder. 

An assumption of our network is that all meshes are in correspondence. While this may seem restrictive, recent work has made use of in-correspondence meshes more convenient. \cite{sharp2020pointtrinet} introduces a neural network for fast computation of triangulations for point clouds. Nonisometric shape matching  \cite{ezuz2019reversible, schonsheck2018nonisometric} can be applied for computing a mapping between meshes, which can put a mesh vertex set in correspondence with a reference mesh. Thus, CFAN-VAE can handle less uniform data with appropriate preprocessing.        

\subsection{Loss Function}

The training loss for our model is given by:
\begin{align}
    & \mathcal{L}  :=  ||\mP - \hat{\mP}||_1  + \sum_{\mu^{(0)}, \mu^{(1)} \in \mu} \left( \lambda_{D} \mathcal{L}_{D}(\mu^{(0)}, \mu^{(1)}) + \lambda_{M}\mathcal{L}_{M}(\mu^{(0)}, \mu^{(1)}) \right) + \lambda_{KL} ||\sigma^2 + \mu^2 - 1 - \log(\sigma^2) ||_1 .
\end{align}
The first term is simply the $L_1$ error of the reconstruction which promotes robustness to vertex outliers. 
$\mathcal{L}_D$ and $\mathcal{L}_M$ are disentanglement and metric penalties that we define shortly. The fourth term of the loss function is the KL-divergence of the latent representation from the unit normal distribution, given by the Bayesian prior assumption \cite{kingma2013auto}.


In CFAN-VAE, geometric disentanglement is a result of separately encoding intrinsic and extrinsic geometric information. We stress that structural conditions prevent guaranteeing complete independence of the separate latent vectors. This is intuitive, considering motion can be restricted by intrinsic properties such as body mass distribution. 

To promote independence, we consider a pair of encoded meshes. We create a latent variable with the conformal latent code of the first mesh, while the normal code is a random convex combination of the two normal codes. We decode and re-encode this new latent variable. We then have an $L_2$ penalty on the resulting change in the conformal code upon re-encoding as it should not be affected by this change in the normal code. We penalize a change in the normal code analogously. This disentanglement penalty, $\mathcal{L_D}$, is formalized below,        
\begin{align}
\mathcal{L}_{D} :=  & \E \sum_{\substack{i, j \text{ in}\\ \{c, n\}}} ||\mu_i^{(0)} - E_i \left( D(\mu_c^{(0)} + \delta_{j, c} \epsilon_c, \mu_n^{(0)} + \delta_{j, n} \epsilon_n) \right)||_2^2,   \\
\epsilon_k & := \alpha (\mu_k^{(1)} - \mu_k^{(0)}), \quad \alpha \in U[0, 1]. 
\end{align}

Inspired by the metric regularization introduced in \cite{cosmo2020limp}, we add a metric penalty, $\mathcal{L}_M$, to the loss function,
\begin{align}
\mathcal{L}_{M} :=  & \E_\alpha || l_\alpha(\phi_c \circ D(\mu^{(0)}), \phi_c \circ D(\mu^{(0)}))  - \phi_c \circ D \left( l_\alpha(\mu^{(0)}, \mu^{(1)})\right) ||_1, \quad \alpha \in U[0, 1],  
\end{align}
where $l_\alpha$ is the $\alpha$ convex combination of its entries. 
This promotes smoothness in the metric of reconstructions. We include an ablation study of these loss terms in Table \ref{tab:loss_ablation} in the appendix. We find that the disentanglement and metric penalties are needed to reduce their respective errors.

\begin{figure*}[tb]
    \centering
    \begin{subfigure}[b]{0.30\textwidth}
    \includegraphics[width=\textwidth]{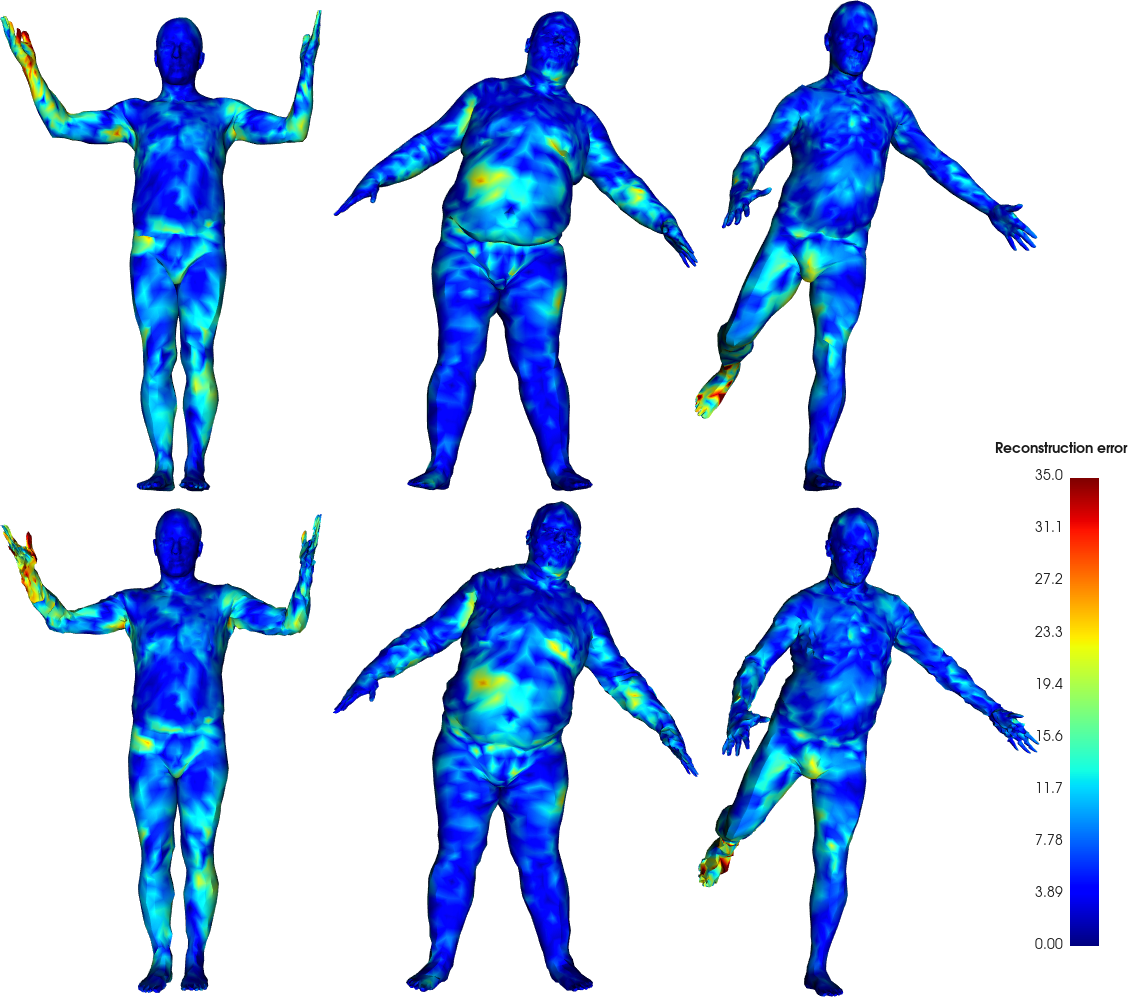}
    \end{subfigure}
    \hfill 
    \begin{subfigure}[b]{0.23\textwidth}
    \includegraphics[width=\textwidth]{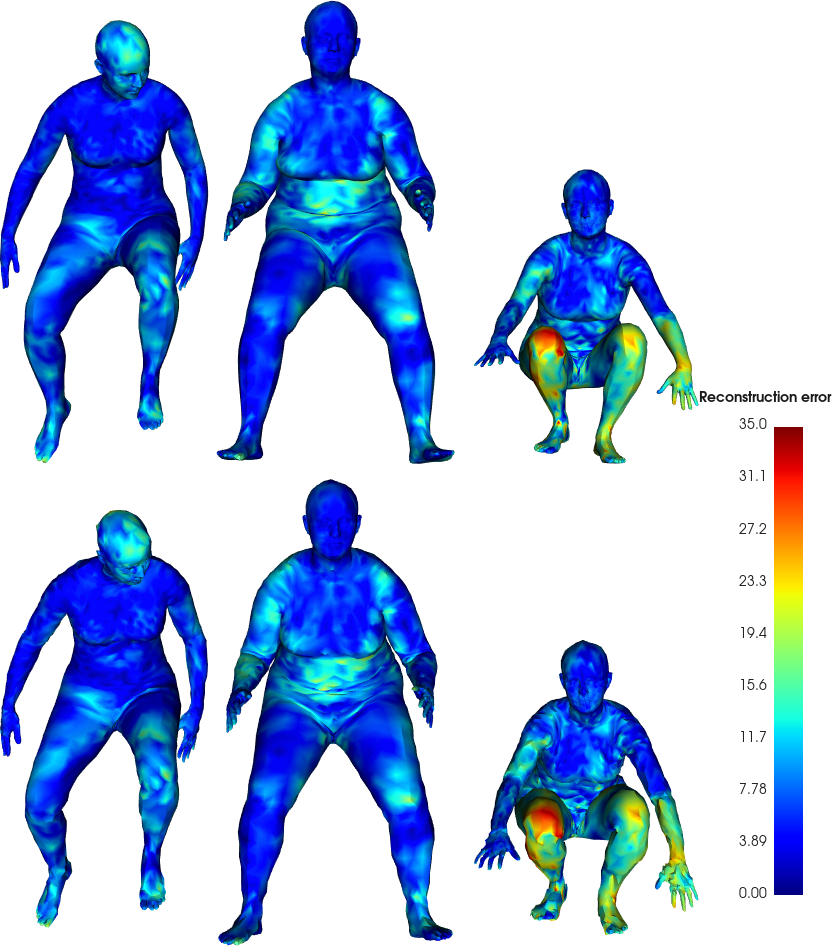}
    \end{subfigure}
    \hfill
    \begin{subfigure}[b]{0.38\textwidth}
    \includegraphics[width=\textwidth]{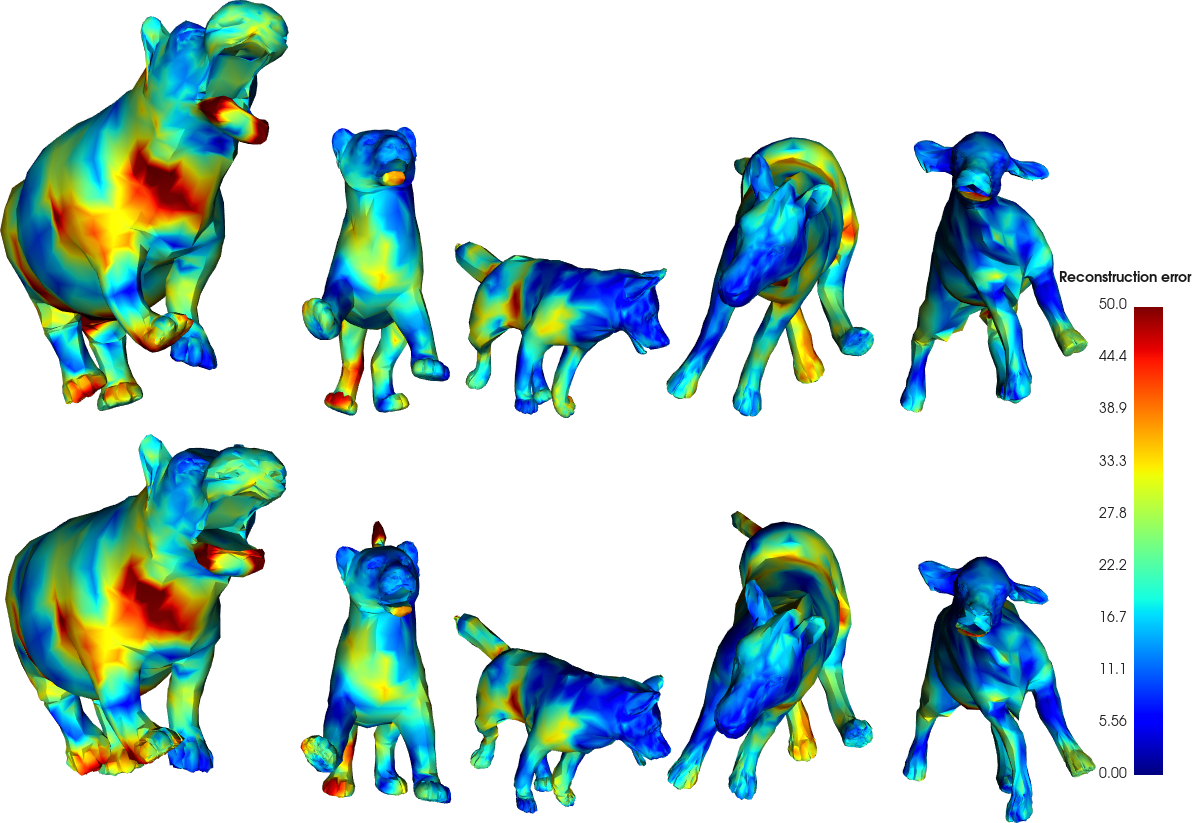}
    \end{subfigure}
    \caption{Reconstructions from the CFAN-VAE models in Figure \ref{fig:interp}. The mesh color represents the pointwise Euclidean error after translation. The top row are the ground truth meshes, and the bottom row are the reconstructions. 
    }
    \label{fig:reconstructions}
\end{figure*}

\section{Applications of CFAN-VAE}

\paragraph{Identity and Pose Transfer}

Interpolating data is a popular task for VAEs. To achieve this, we encode two meshes, linearly interpolate between their latent code, and decode this interpolation. A similar task is style transfer, where we smoothly transfer the pose or identity of one mesh onto another while keeping the other property fixed. Geometric disentanglement provides a simple way to do this. 

Transferring identity is achieved via interpolating the conformal latent codes, $\vz_{c}$, while keeping the normal latent codes, $\vz_{n}$, fixed. Formally, given two surfaces with latent codes, $(\vz_c^{(0)}, \vz_n^{(0)})$ and $(\vz_c^{(1)}, \vz_n^{(1)})$,
\begin{equation}
    \hat{\vp}_c(\alpha) = D((1 - \alpha) \vz_c^{(0)} + \alpha \vz_c^{(1)}, \vz_n^{(0)}), \quad \alpha \in [0, 1]
\end{equation}
is a path that smoothly transfers the identity of the second surface onto the first. Transferring pose is analogous. 

\paragraph{Identity and Pose Generation}
Another typical task for VAEs is the generation of new surfaces via decoding samples from the latent space. Since the goal of CFAN-VAE is to achieve geometric disentanglement, we expect to randomly generate surfaces of different identity with the same pose and vice versa. This is accomplished through separate samplings of $\vz_{c}$ and $\vz_{n}$. That is,
\begin{equation}
    \hat{\vp}_c^{(i)} = D(\vz_c^{(i)}, \vz_n^{(0)}), \quad \vz_c^{(i)} \sim \mZ_c, \text{ for } i \in I
\end{equation}
corresponds to generated meshes of different identities in the same pose. Generating with identity fixed is analogous.

\paragraph{Latent Space Registration}
A natural question is to what extent CFAN-VAE can provide insight on the entanglement in xyz-VAEs and disentangle them. It can further lead to greater interpretability of the xyz-VAE latent space, though it requires a trained CFAN-VAE. We explore this through Orthogonal Procrustes (OP) analysis~\cite{schonemann1966generalized} and find that disentanglement is indeed possible without label information. 

Here, we use OP analysis to register two different latent representations of the same dataset, $\mZ_{x} \in \mathbb{R}^{d \times \ell}$ encoded by a xyz-VAE and $\mZ_{c, n} \in \mathbb{R}^{d \times \ell }$ encoded by a CFAN-VAE, where $d$ and $\ell$ denote the number of data samples and the latent dimension respectively. Then the OP problem is:
\begin{equation}\label{eq:ortho_proc}
    \mR^* = \argmin_{\mR \in SO(d)} ||\mZ_{x} \mR - \mZ_{c, n}||_F^2 = \mU \mV^T \quad 
\end{equation}
where $\mU, \mV$ are provided from a singular value decomposition $ \mZ_x^T \mZ_{c, n} = \mU \mSigma \mV^T$. 
Since $\mZ_{c, n}$ is the concatenation of the conformal and normal latent codes, we write $\mR^{*} = [\mR^{*}_c | \mR^{*}_n]$ as two matrices concatenated along the column dimension. This allows us to define latent subspaces:
\begin{equation}
\label{eqn:LatentVarReg}
    \mZ_{x, c} = \mZ_x \mR^{*}_c, \mZ_{x, n} = \mZ_x \mR^{*}_n, \text{ s.t. } \mR^{*} = [\mR^{*}_{c} | \mR^{*}_{n}].
\end{equation}
Thus, we identify corresponding conformal and normal latent codes in the xyz-VAE latent space: $\vz_{x, c}$ and $\vz_{x, n}$. In section \ref{sec:results}, we demonstrate these registered latent subspaces exhibit significant geometric disentanglement. 



\begin{figure*}[tb]
    \centering
    \begin{subfigure}[b]{0.27\textwidth}
    \includegraphics[width=\textwidth]{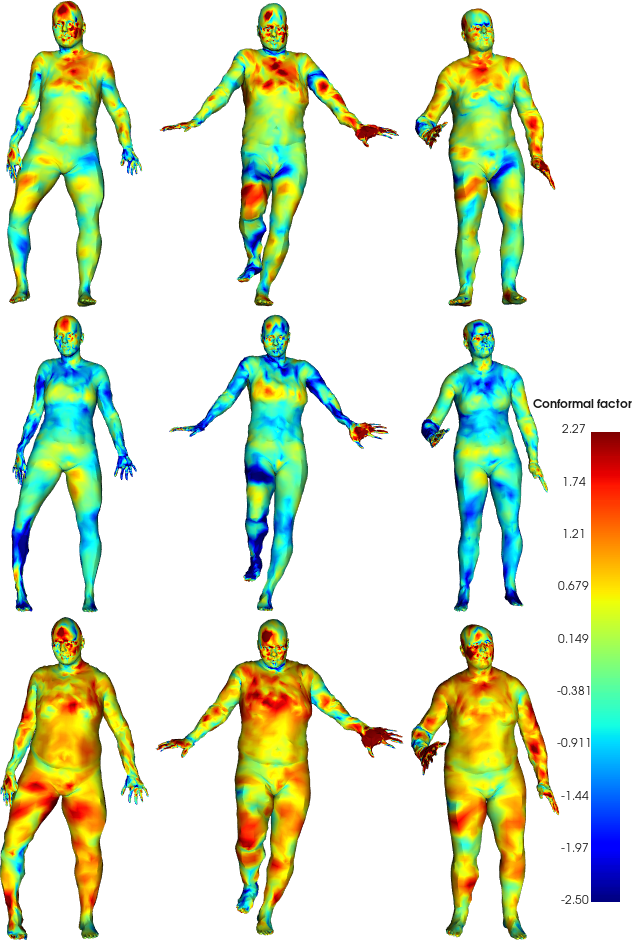}
    \end{subfigure}
    \hfill 
    \begin{subfigure}[b]{0.24\textwidth}
    \includegraphics[width=\textwidth]{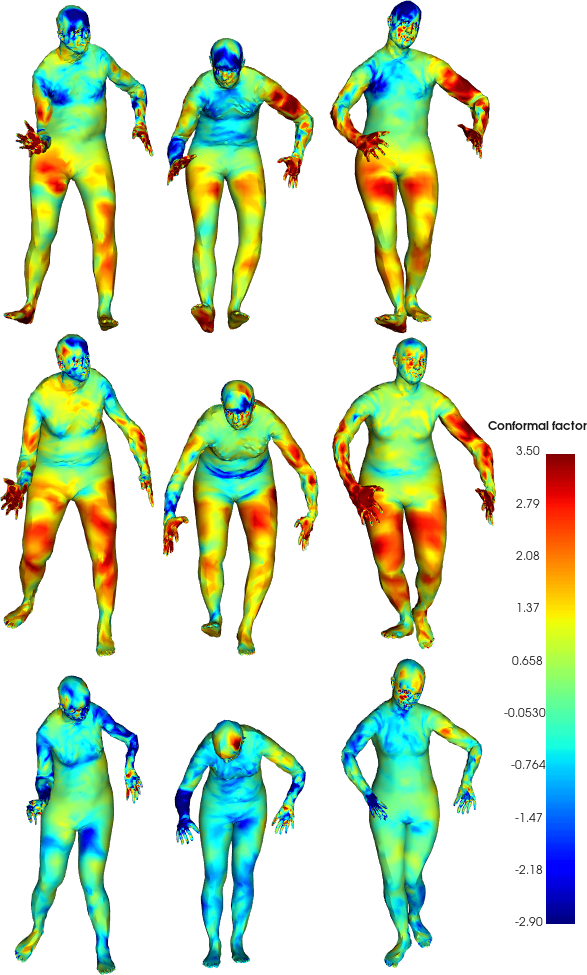}
    \end{subfigure}
    \hfill
\begin{subfigure}[b]{0.26\textwidth}
    \includegraphics[width=\textwidth]{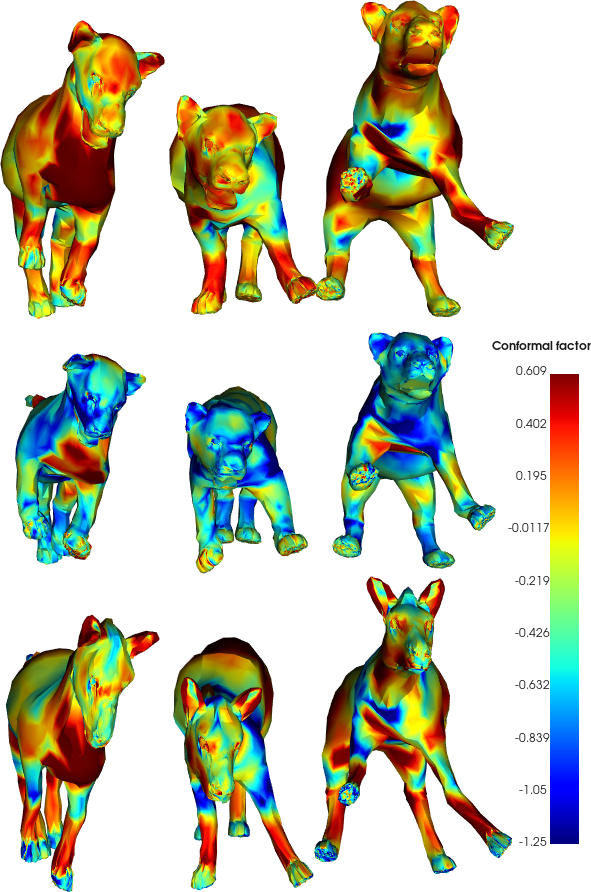}
    \end{subfigure}
    \caption{Surfaces generated by sampling the CFAN-VAE latent space from a multivariate Gaussian distribution. 
    The horizontal and vertical axes correspond to $\vz_{n}$ and $\vz_{c}$ respectively. CFAN-VAE reliably generates meshes in a specific pose or with nearly the same identity, showing geometric disentanglement. Color denotes conformal factor.}
    \label{fig:cfan_noise}
\end{figure*}

\section{Numerical Experiments}
We demonstrate the efficacy of the CFAN-VAE through numerical experiments including reconstruction, interpolation, generation, and latent space analysis. We also show how a CFAN-VAE can be used to disentangle the latent space of a xyz-VAE. All numerical experiments were conducted with NVIDIA 2080 Ti GPUs. 

\paragraph{Datasets}
We consider three datasets in our experiments. This includes DFAUST, a real-world motion-capture dataset of 10 people performing several poses \cite{dfaust:CVPR:2017}. We randomly split the dataset into a training/validation/test set of 35,720/500/5,000 meshes respectively. 

In addition, we generated a synthetic dataset of animals referred to as SMAL following \cite{Zuffi:CVPR:2017}, based on the SMPL model introduced in \cite{Loper2015SMPLAS}. Similary, we formed a synthetic dataset of humans referred to as SURREAL from \cite{varol17_surreal}. These datasets are created following \cite{aumentadoarmstrong2019geometric}. For SURREAL, we add small normal noise with $\sigma = 0.2$ to the intrinsic shape parameters $\beta$ to eliminate any isometry in the dataset. These two datasets provide challenging environments where identity-supervised methods fail. 
Additional details on the generation can be found in Appendix \ref{subsec:app_datasets}. 

\paragraph{Architecture Hyperparameters} 
Each VAE contains 5 layer encoders and decoder with details described in Appendix \ref{subsec:app_architecture}. For the PTC kernels, we use a 13 point stencil with the support of the kernel about a point being its 9 nearest geodesic neighbors. The size of the conformal/normal latent vectors are (32, 32) for DFAUST and (8, 32) for SMAL. For SURREAL, we train models with several latent dimension sizes for ablation studies found in Table \ref{tab:latent_ablation} in the Appendix. For visualization, we use models with a latent dimension of (32, 32). For corresponding xyz-VAEs, the latent dimension is the sum of the two latent dimensions.  

For comparisons to GDVAE, we use the available dataset-specific pretrained models available in \cite{aumentadoarmstrong2019geometric}. The latent dimension of the GDVAE models are $(5, 15)$ and $(5, 11)$ for SURREAL and SMAL respectively. When comparing to CFAN-VAE models for SURREAL, we choose a corresponding latent dimension of $(5, 15)$. 


We use batch normalization (BN) layers in the encoder. 
When training with the disentanglement penalty $\mathcal{L}_D$, we freeze the BN layers during reencoding. The activation functions are ReLU and ELU for the encoders and decoder. We use the AdamW optimizer \cite{loshchilov2017decoupled} with a learning rate of 1E-3 and a weight decay of 5E-5. The value of $\lambda_{KL}$ is set at 1E-4, while $\lambda_{D}$ and $\lambda_M$ are 5E-2. Each model is trained for 300 epochs with a batch size of 32.  We train two instances of each network using different random seeds.

\begin{table*}[tb]
    \centering 
    \begin{tabular}{@{}lcrrcrr@{}}
    \toprule
     & \phantom{abc} & \multicolumn{2}{c}{SURREAL} & \phantom{ab} & \multicolumn{2}{c}{SMAL} \\
     \cmidrule{3-4} \cmidrule{6-7}
     Networks && Error(mm) & \#Param.(M) && Error(mm) & \#Param.(M) \\
    \midrule
    Distance between sources && $76.0 \pm 1.8$ & --- && $313.9 \pm 9.7$ & --- \\
    GDVAE \cite{aumentadoarmstrong2019geometric} && $31.6 \pm 0.6$ & $21.30$ && $96.3 \pm 1.9$ & $21.30$ \\
    Registered XYZ-VAE && $28.9 \pm 0.7$ & $1.68$ && $56.3 \pm 1.6$ & $1.80$\\ 
    CFAN-VAE && $\mathbf{26.8 \pm 0.8}$ & $1.75$ && $\mathbf{35.5 \pm 0.8}$ & $2.15$  \\
    \bottomrule
    \end{tabular}
    \caption{Chamfer distance of the mesh generated in pose and identity transfer tasks to the target mesh. The generated mesh is decoded from swapping the latent codes of the source meshes, $(\vz_c^{(1)}, \vz_n^{(1)})$ and $(\vz_c^{(2)}, \vz_n^{(2)})$. This reported distance is the mean over 256 transfer pairs with standard error. CFAN-VAE is clearly best at the transfer task.}
    \label{tab:transfer}
\end{table*}


\begin{figure*}[tb]
\centering
\begin{subfigure}[b]{.28\textwidth}
  \includegraphics[width=\textwidth]{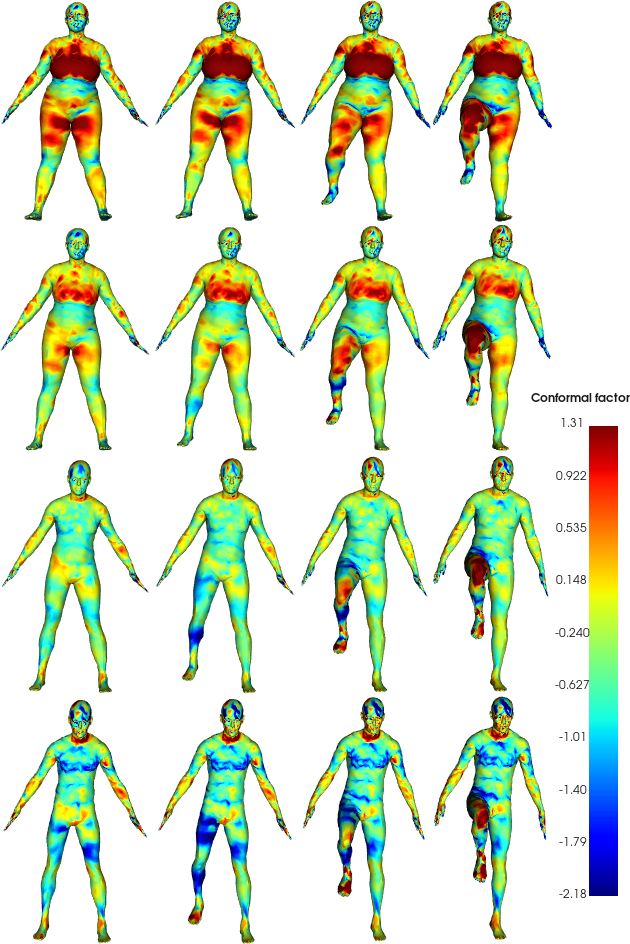}
\end{subfigure}
\hfill
    \begin{subfigure}[b]{.25\textwidth}
    \includegraphics[width=\textwidth]{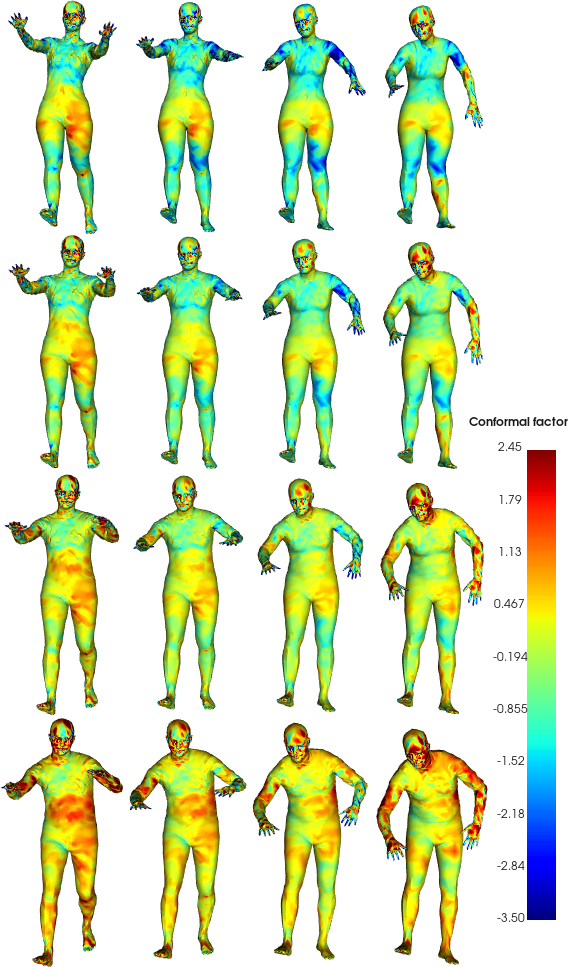}
    \end{subfigure}
\hfill
\begin{subfigure}[b]{.38\textwidth}
\includegraphics[width=\textwidth]{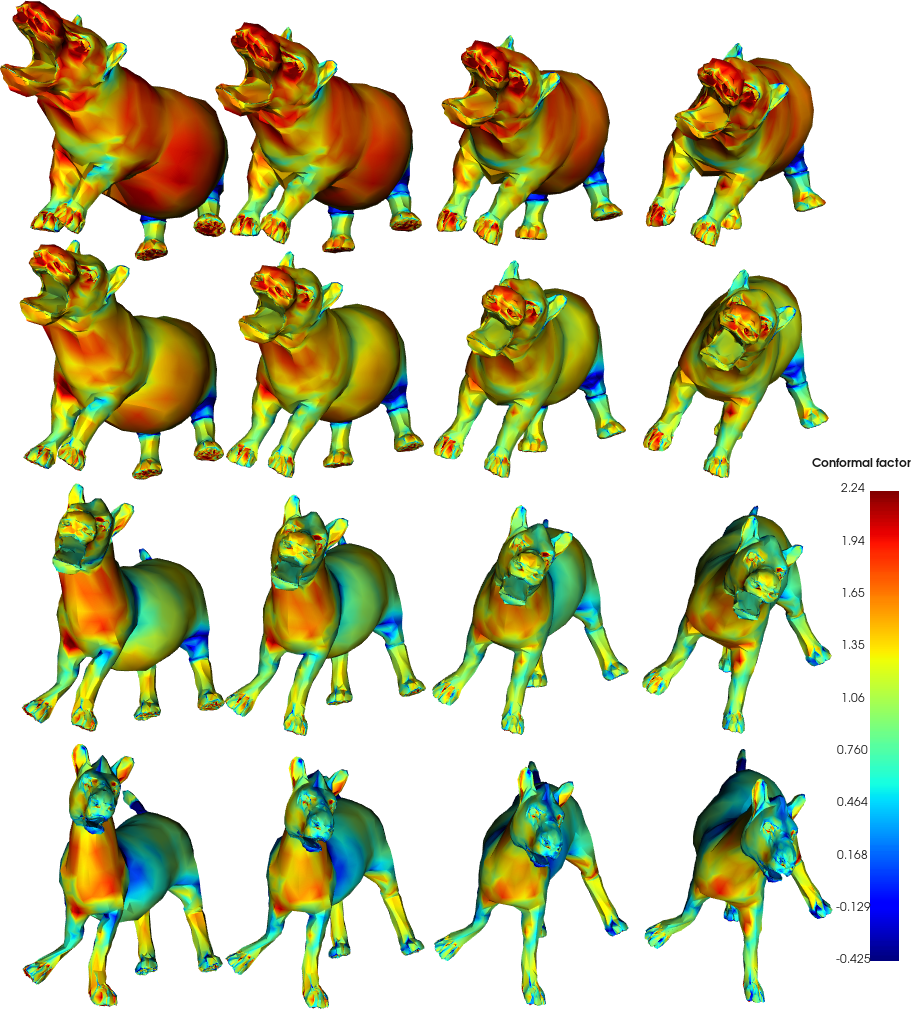}
\end{subfigure}
\caption{
Interpolations for xyz-VAEs, using $\vz_{x, n}$ and $\vz_{x, c}$ determined by the OP registration to a CFAN-VAE. Geometric disentanglement in xyz-VAEs is detected, but requires registration. Color denotes conformal factor.
}
\label{fig:interp_xyz}
\end{figure*}

\subsection{Results} \label{sec:results}

\paragraph{Surface Reconstruction}
We verify that CFAN-VAE is able to accurately reconstruct surfaces.
Figure \ref{fig:reconstructions} displays high quality reconstructions for each dataset using CFAN-VAEs for DFAUST, SURREAL, and SMAL. Table \ref{tab:error} in the appendix contains quantitative measurements related to the performance for each CFAN-VAE. The generalization error is the average $L_2$ vertex error on the reconstruction. Clearly CFAN-VAE can reconstruct high quality meshes.



\paragraph{Surface Interpolation}

Figure \ref{fig:interp} displays examples of disentangled interpolations for each dataset using CFAN-VAE. In this figure, the vertical and horizontal axis represents conformal and normal interpolations respectively. The color map denotes the pointwise normalized conformal factor of the reconstructions. These results verify that CFAN-VAE produces meaningful geometric disentanglement between intrinsic and extrinsic information. Thus, it enjoys flexibility to generate high quality meshes preserving either identity or pose. 
Notice, the colormaps and position are almost identical for fixed $\vz_c$ and $\vz_n$ respectively. We include additional examples in Appendix \ref{sec:add_figures}.

We compare the performance of CFAN-VAE to GDVAE on transfer tasks. For SURREAL and SMAL, we can create ground-truth mesh solutions to the transfer task by exchanging the appropriate SMPL parameters of the source meshes. Then we compare the Chamfer distance of the network generated mesh, with appropriate latent code exchanged, to the SMPL generated ground truth. These results are summarized in Table \ref{tab:transfer}. As GDVAE computes Chamfer distance to a uniformly sampled set of $1500$/$2000$ points for SMAL/SURREAL, when computing the distance for CFAN-VAE, we downsample the points to the first level of encoding to produce a comparable $2048$ uniformly sampled points. It is clear that CFAN-VAE outperforms GDVAE, particularly on the SMAL dataset. This nice property of the CFAN-VAE latent space enables flexible control of the geometry of generated surfaces. We stress that these two datasets lack isometric pairs, so other methods \cite{cosmo2020limp, zhou2020unsupervised} are not of use for comparison.   

\paragraph{Surface Generation}

Figure \ref{fig:cfan_noise} plots random generations where latent codes are sampled from a multivariate Gaussian. This Gaussian, $\mathcal{N}(\vmu, \mSigma)$, is a product of marginals fit to the embedding of the test set for each latent variable. We sample from $\mathcal{N}(\vmu, 0.8 \mSigma)$. Here the normal latent code and conformal code is fixed along the vertical axis and horizontal axis respectively. Clearly, pose and identity are successfully fixed. The generated meshes are of comparable visual quality to the reconstructions from the test set. Then the latent space of CFAE-VAE allows for high-quality generation while providing interpretability of geometric features.

\begin{table*}[tb]
    \centering 
    \begin{tabular}{@{}lcccccrrcrr@{}}
    \toprule
     && \phantom{a} & \multicolumn{2}{c}{Desired} & \phantom{a} & \multicolumn{2}{c}{SURREAL} & \phantom{ab} & \multicolumn{2}{c}{SMAL} \\
     \cmidrule{7-8} \cmidrule{10-11}
     Networks & Parameter && \multicolumn{2}{c}{Magnitude} && $\vz_c$ & $\vz_n$ && $\vz_c$ & $\vz_n$ \\
    \midrule
    \multirow{2}{*}{GDVAE \cite{aumentadoarmstrong2019geometric}}  & $\beta$ && $\downarrow$ & $\uparrow$ && $4.42 \pm 0.03$ & $4.44 \pm 0.03$ && $2.40 \pm 0.02$ & $3.48 \pm 0.02$ \\
    & $\theta$ && $\uparrow$ & $\downarrow$ && $4.21 \pm 0.04$ & $4.11 \pm 0.04$ && $2.72 \pm 0.01$ & $2.71 \pm 0.01$\\
    \cmidrule{1-2}
    Registered & $\beta$ && &&& $4.19 \pm 0.02$ & $4.38 \pm 0.02$ && $1.84 \pm 0.01$ & $3.38 \pm 0.02$\\ 
    XYZ-VAE & $\theta$ && &&& $3.69 \pm 0.03$ & $2.83 \pm 0.04$ && $2.72 \pm 0.01$ & $2.64 \pm 0.01$ \\
    \cmidrule{1-2}
    \multirow{2}{*}{CFAN-VAE} & $\beta$ && &&& $\mathbf{4.05 \pm 0.02}$ & $4.38 \pm 0.02$ && $\mathbf{1.73 \pm 0.01}$ & $3.73 \pm 0.02$  \\
    & $\theta$ && &&& $3.69 \pm 0.03$ & $\mathbf{2.81 \pm 0.04}$ && $2.74 \pm 0.01$ & $\mathbf{2.63 \pm 0.01}$\\
    \bottomrule
    \end{tabular}
    \caption{Metrics evaluating disentanglement in the latent space learned by the networks. For each mesh embedding, we search for its nearest neighbor in either the intrinsic or extrinsic latent space, $\vz_c$ / $\vz_n$. We report the norm of the difference in the identity and pose SMPL parameters, $\beta$ and $\theta$, of the embedded mesh and its neighbor. We expect the shortest distances to be with respect to $\beta$ for $\vz_c$ neighbors and $\theta$ for $\vz_n$ neighbors. We report mean distance with standard error on the test set.}
    \label{tab:NNdist}
\end{table*}

\begin{figure*}[tb]
    \centering
    \begin{subfigure}[b]{0.44\textwidth}
    \includegraphics[width=\textwidth]{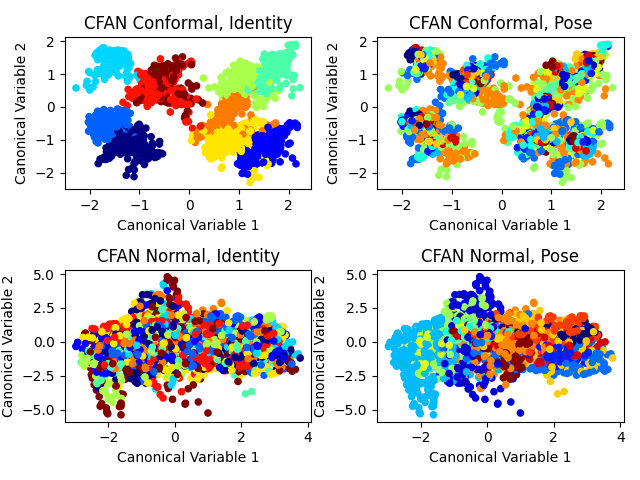}
    \caption{CFAN-VAE}
    \label{subfig:cca_CVAE}
    \end{subfigure}\hspace{0.8cm}
    \begin{subfigure}[b]{0.44\textwidth}
    \includegraphics[width=\textwidth]{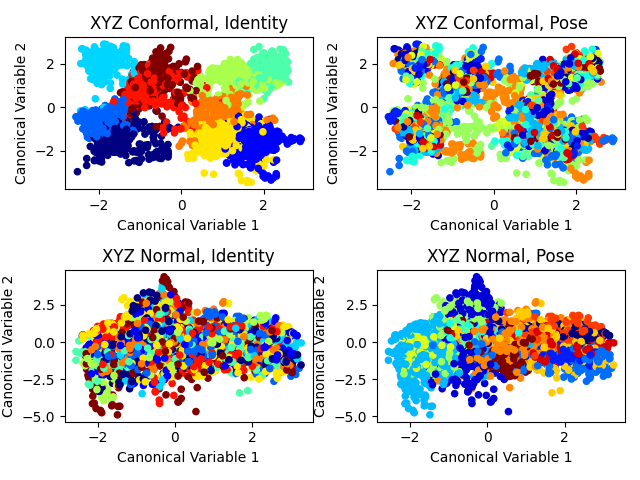}
    \caption{xyz-VAE}
    \label{subfig:cca_MeshVAE}
    \end{subfigure}
    \caption{2D canonical correlation analysis (CCA) embeddings of registered latent spaces. The networks are a CFAN-VAE/xyz-VAE trained on DFAUST. CCA is performed for the pairs, $\vz_{c}$/$\vz_{x,c}$ (top rows) and $\vz_{n}$/$\vz_{x, n}$ (bottom rows). The color in the left/right columns denotes the identity/pose associated with the embedded mesh. 
    }
    \label{fig:cca_plots}
\end{figure*}

\paragraph{Latent Registration}

We verify the ability to disentangle xyz-VAE latent space using CFAN-VAE and OP registration in Figure \ref{fig:interp_xyz}. We display the interpolations for each dataset using the registered xyz-VAE latent space based on \eqref{eqn:LatentVarReg}. There is clear geometric disentanglement, although still some entanglement. For example, in the SMAL registered xyz-VAE interpolation, the head of the top row hippo and bottom row horse are not aligned though they make the same general motion. Notably, this is not a problem for the same interpolation using the CFAN-VAE in Figure \ref{fig:interp}. In Table \ref{tab:transfer}, we quantitatively compare the performance of the registered xyz-VAE on identity and pose transfer tasks. This network outperforms GDVAE, while its performance is less than the corresponding CFAN-VAE. 

It is significant that CFAN-VAE can identify geometric disentanglement in a network trained without disentanglement as a goal. The latent axes learned from a xyz-VAE entangle intrinsic and extrinsic geometry, but we are able to use the CFAN-VAE to identify disentangled variables.

\paragraph{Latent Space Analysis}
To better assess the geometric disentanglement in our networks, we analyze the latent spaces of the CFAN-VAEs and registered xyz-VAEs. We quantitatively measure the disentanglement in latent space in the manner of \cite{zhou2020unsupervised}. First, we embed the test sets for SURREAL and SMAL in the latent space. Given a mesh embedding, we find its nearest neighbor in either the intrinsic or extrinsic latent spaces, $\mZ_c$ or $\mZ_n$. As these meshes were created via SMPL shape and pose parameters, $\beta$ and $\theta$, we measure the distance between these parameters of the embedded mesh and its neighbor. These distances are summarized in Table \ref{tab:NNdist}. If the network latent spaces truly possess disentanglement, we expect the $\vz_c$ neighbor to be closer in $\beta$ distance and the $\vz_n$ neighbor to be closer in $\theta$ distance. 

We see that CFAN-VAE performs the best of the three networks with GDVAE performing worse than the other two. For SMAL, the latent dimension of CFAN-VAE, $(8, 32)$, was considerably larger than that of GDVAE, $(5, 11)$. Thus, we use the $(5, 11)$ dimension PCA embeddings of the CFAN-VAE latent spaces for comparison. We believe this comparison is  fair given the large disparity in number of trainable parameters as seen in Table \ref{tab:transfer}.   

Additionally, we offer a visualization of this latent geometric disentanglement. While it is not apt for quantitative analysis of transfer tasks, DFAUST possesses identity and action labels. In Figure \ref{fig:cca_plots}, we visualize identity and action clustering in the latent space of the CFAN-VAE and xyz-VAE using canonical correlation analysis (CCA). For either the conformal or normal subspaces in CFAN-VAE and xyz-VAE, this method finds the two linear transformations that whiten the individual data embeddings, while maximizing their cross-correlation. We refer to \cite{hardoon2004canonical} for details on CCA. 

We see that expections of clustering are met in this visualization; identity clusters only exist in the conformal embeddings and action clusters only exist in the normal embeddings. These results strongly support the existence of geometric disentanglement in CFAN-VAEs.
Additionally, it is visually clear that the CFAN-VAE and xyz-VAE embeddings are incredibly similar. This implies that the latent spaces of these different architectures are highly correlated. 


\section{Conclusion}
We propose a novel architecture, CFAN-VAE, for unsupervised geometric disentanglement in mesh convolutional autoencoders. This is accomplished by utilizing the conformal factor and normal (CFAN) feature to separate intrinsic and extrinsic geometry. There is clearly strong geometric disentanglement in CFAN-VAEs. Additionally, we detect geometric disentanglement in xyz-VAEs by registering their latent spaces to those of CFAN-VAEs. We achieve state-of-the-art performance on transfer tasks and disentanglement compared to prior work. The continued integration of geometric theory into neural network architectures can only lead to more interesting results in the future.  


%
%

{
\bibliographystyle{ieeetr}
\bibliography{main}
}

\clearpage

\appendix

\section*{Appendix}

\begin{table*}[tb]
\begin{center}
    \begin{tabular}{@{} l r r c r r r@{}}
    \toprule
     \multirow{2}{*}{Dataset} & \multirow{2}{*}{\#$\vz_c$} & \multirow{2}{*}{\#$\vz_n$} &  \phantom{ab} & \multicolumn{3}{c}{CFAN-VAE}  \\
     \cmidrule{5-7}
     &&& & $L_1$ (E-1) & $\KL$ (E2) & Error (mm) \\
     \midrule
     \multirow{1}{*}{DFAUST} & 
     32 & 32
     && 1.46 $\pm$ 0.01 & 2.22 $\pm$ 0.04 & 8.37 $\pm$ 0.06 \\
     
     \cmidrule{1-1}
     \multirow{4}{*}{SURREAL}     
     & 5 & 15 
     && 1.56 $\pm$ 0.01 & 0.97 $\pm$ 0.05 & 11.69 $\pm$ 0.14 \\
     & 8 & 8 
     && 2.43 $\pm$ 0.01 & 0.79 $\pm$ 0.01 & 18.24 $\pm$ 0.07 \\
     & 16 & 16 
     && 1.27 $\pm$ 0.02 & 1.59 $\pm$ 0.01 & 9.51 $\pm$ 0.35  \\
     & 32 & 32 
     && 1.14 $\pm$ 0.01 & 2.66 $\pm$ 0.02 & 8.49 $\pm$ 0.72 \\
    
     \cmidrule{1-1}
     \multirow{1}{*}{SMAL}     
     & 8 & 32 
     && 1.33 $\pm$ 0.01 & 1.74 $\pm$ 0.02 & 25.94 $\pm$ 0.02 \\
     \bottomrule
    \end{tabular}
\end{center}
\caption{The $L_1$ reconstruction loss, KL divergence loss, and the generalization error (mm) on the test set for each CFAN-VAE model. Generalization error is the average Euclidean error of reconstructed vertices. Mean and standard deviation are reported.  
}
\label{tab:error}
\end{table*}

\begin{table*}[tb]
    \centering
    \begin{tabular}{@{} l r r r r c r c r r c r r@{}}
    \toprule
    Loss & $L_1$ & $\mathcal{L}_D$ & $\mathcal{L}_M$ & Error && Transfer && \multicolumn{2}{c}{$\beta$} && \multicolumn{2}{c}{$\theta$} \\
    \cmidrule{9-10} \cmidrule{12-13}
    function & (E-1) & (E-2) & (E-1) & (mm) & \phantom{a} & Err. (mm) & \phantom{a} &  $\vz_c$ &  $\vz_n$ && $\vz_c$ & $\vz_n$ \\
    \midrule
    $L_1$ loss     & 1.16 & 442.00 & 12.30 & 8.67 && 35.5 && 4.21 & 4.43 && 3.51 & 2.85\\
    \quad Add: $\KL$     & 1.15 & 478.70 & 12.33 & 8.63 && 35.5 && 4.19 & 4.42 && 3.52 & 2.85 \\
    \quad Add: $\mathcal{L}_D$ & 1.18 & 2.27 & 12.38 & 8.86 && 32.1 && 4.20 & 4.44 && 3.49 & 2.86 \\
    \quad Add: $\mathcal{L}_M$ & 1.27 & 2.16 & 6.71 & 9.51 && 33.1 && 4.23 & 4.43 && 3.58 & 2.84\\
    \bottomrule
    \end{tabular}
    \caption{Ablation study concerning the loss function on CFAN-VAE trained on SURREAL with latent dimensions $(16, 16)$. The disentanglement and metric penalties have a small negative impact on generalization error. However, the disentanglement penalty is clearly important for constraining disentanglement error, seen in $\mathcal{L}_D$ as well as the transfer task error. The metric penalty reduces error in the reconstructed metric, improving qualitative appearance of the meshes.}
    \label{tab:loss_ablation}
\end{table*}

\begin{table*}[tb]
    \centering
    \begin{tabular}{@{} r r c r r c r c r r c r r c r@{}}
    \toprule
    && \phantom{a} & $\mathcal{L}_D$ & $\mathcal{L}_M$ & \phantom{a} & Transfer & \phantom{a}  & \multicolumn{2}{c}{$\beta$} & \phantom{a} & \multicolumn{2}{c}{$\theta$} & \phantom{a} & \#Param  \\
    \cmidrule{9-10} \cmidrule{12-13}
    \#$\vz_c$ & \#$\vz_n$ &&  (E-2) &  (E-1) && Err. (mm) &&  $\vz_c$ & $\vz_n$ && $\vz_c$ & $\vz_n$ && (K) \\
    \midrule
    8 & 8 && 2.21 & 7.43 && 45.1 && 4.32 & 4.43 && 3.50 & 3.02 && 456.2 \\
    16 & 16 && 2.16 & 6.71 && 33.1 && 4.23 & 4.43 && 3.58 & 2.84  && 497.2 \\
    32 & 32 && 4.32 & 6.91 && 31.3 && 4.21 & 4.43 && 3.51 & 2.86 && 579.2 \\
    \bottomrule
    \end{tabular}
    \caption{Ablation study regarding the latent dimension choice of CFAN-VAE trained on SURREAL. We see that performance on the transfer tasks improve with greater latent dimension, whereas the disentanglement in the latent space measured by SMPL parameter distance is mostly the same. The increased performance on the transfer task is likely due to the increased parameterization of the network.}
    \label{tab:latent_ablation}
\end{table*}

\begin{table*}[tb]
    \centering
    \begin{tabular}{@{} l l c r r r @{}}
    \toprule
    \multicolumn{2}{c}{Noise} \\
    \cmidrule{1-2}
    Trained & Evaluated & \phantom{a} & $L_1$ (E-1) & $\mathcal{L}_M$ (E-1) & Error (mm) \\
    \midrule 
    No & No  && 1.27 & 6.71 & 9.51  \\
    Yes & No && 1.49 & 7.96 & 11.13 \\
    \cmidrule{1-2}
    No & Yes && 1.83 & 9.65 & 13.66 \\
    Yes & Yes && 1.51 & 9.09 & 11.29 \\
     \bottomrule
    \end{tabular}
    \caption{An analysis on the effect of noise on a CFAN-VAE trained on SURREAL with latent dimensions $(16, 16)$. In training and/or evaluation, noise sampled from $\mathcal{N}(0, 1mm)$ is added to each vertex, which translates to large noise on the surface normals. We see that it is possible to train a CFAN-VAE with noise so that it is robust to noise.}
    \label{tab:noise_ablation}
\end{table*}

\section{Background}
\label{sec:background}
In this section we review background on variational autoencoders and geometric deep learning.

\subsection{Variational Autoencoders}
\label{subsec:vae_background}
The variational autoencoder (VAE) is a deep generative model introduced in \cite{kingma2013auto}. Assume the input into the generative model is $x \in \R^n$. Then the VAE consists of a probabilistic encoder, $q_\theta(\rz | \rx_i)$ , and a probabilistic decoder, $p_\phi(\rx_i | \rz)$. Here, we have the latent variable, $\rz \sim q_\theta(\rz | \rx_i)$, as well as the reconstruction $\hat{x}$, where $\hat{\rx_i} \sim p_\phi(\rx_i | \rz)$. We also assume a prior distribution, $p(\rz)$, on the latent variable. 

The VAE loss is described in \eqref{eq:vae} below. This loss is composed of the negative log likelihood of the reconstruction and the KL divergence of the variational posterior from the prior on the latent variable. 
\begin{align}\label{eq:vae}
    \mathcal{L}(\theta, \phi) := &\sum_{i=1}^n - \E_{\rz \sim q_\theta(\rz | \rx_i)}[\log p_\phi(\hat{\rx_i} | \rz)]   + \KL(q_\theta(\rz | \rx_i) || p(\rz))
\end{align}
In practice, a typical prior distribution for $\rz_i$ is the unit Gaussian, $\mathcal{N}(0, 1)$, and all latent variables are assumed to be independent. 

The benefit of using a VAE with a Gaussian prior is that each point on the data manifold is encoded into a Gaussian distribution in the latent space as opposed to an individual point. Then loosely speaking, the distribution on the latent space embedding follows a Gaussian mixture model. This promotes continuity in the latent space embedding, which is important for interpolating between data. Additionally, with each embedded point corresponding to something near a unit Gaussian, data generation can be reliably performed by decoding a sample from the unit Gaussian distribution. 

\subsection{Parallel Transport Networks}
\label{subsec:geodeeplearn}

Geometric deep learning describes a set of methods used to generalize neural networks to better represent non-Euclidean data (such as manifolds, meshes, and graphs) and incorporate geometric information of this data into the learning process. One common challenge is to define a convolutional operation which is both compatible with the domain and useful in CNNs. This can be challenging because these domains do not have the regularity of Euclidean spaces in which convolution can be computed as a sliding window. 

In this work we use Parallel Transport Convolution (PTC), introduced in \cite{schonsheck2018parallel}, to construct convolutional layers in our networks. PTC defines a convolution with a compactly supported kernel in $\mathbb{R}^2$ with a signal supported on a manifold. This is conducted by using parallel transportation, to define a translation-like operation on the tangent space, and the exponential map, to transfer kernels between the tangent space and the manifold. Importantly, all necessary geometric information can be precomputed so that the PTC layer can be implemented as a sparse matrix multiplication with computational costs on the same order as Euclidean convolution (see \ref{subsec:geodeeplearn} for details). 

\subsubsection{Parallel Transport Networks}
\label{subsubsec:PTC}
The parallel transport convolution network (PTCNet) introduced the eponymous convolution in \cite{schonsheck2018parallel}. Previous local patch-based methods in \cite{boscaini2016learning, Monti_2017} defined filters locally without making it apparent how to consistently translate the kernel at vertex $v_i$ to $v_j$. This means that the local frame of the kernel corresponds to global vector fields with many discontinuities. The discontinuities are precisely what PTCNet rectifies.

If we denote $(P_{p_0}^{p_i})(\cdot): T\M_{p_0} \rightarrow T\M_{p_i}$ as the parallel transportation of a vector from the tangent plane at $p_0$ to the tangent plane at $p_i$ then the discretized PTC of a compactly supported filter $k(x_0,\cdot)$ with a signal $f: \M \rightarrow \mathbb{R}$ is given by:

\begin{align}
\label{eq:ptc_conv}
    f*_{\M} k = & \int_{\mathcal{M}} k(p_0, \exp_{p_0} \circ (P_{p_0}^{p_i})^{-1} \circ \exp_{p_i}^{-1}(p_i))  \cdot f(x) \mathrm{dvol}(x) \\
    \approx & \sum_{j \in \mathcal{N}_i} k(p_0, \exp_{p_0} \circ (P_{p_0}^{p_i})^{-1} \circ \exp_{p_i}^{-1}(p_i)) m(p_i) f_i, 
    \notag
\end{align}
where $\exp_{p_0}$ is the exponential map at $p_0$ and $m(p_i)$ is the local mass element. In discretization, the kernel at $p_0$ is mapped to a template points sampled in the tangent plane $T_{p_0}\mathcal{M}$, transported to the tangent plane at the relevant vertex $v_i$, and then mapped onto a local neighborhood about $p_i$ on the surface $\mathcal{M}$. Practically, the kernel $k$ is chosen to be a linear interpolate of a set of fixed stencil points. Then a PTC network layer can be written as the form,
\begin{equation}
    \sigma(f *_{\M} k) = \sigma( \mW \mF \mM f)  
\end{equation}
where $\mW \in \mathbb{R}^{F' x K}$ are the (learnable) convolutional weights, $\mF$ is the precomputed (fixed) interpolation weights, and $\mM$ is the mass matrix associated with the mesh. Additionally, $\sigma$ denotes the nonlinear activation function for the convolutional layer.

The operator $P_{p_0}^{p_i}$ requires a pre-defined vector field to produce local frames on the manifold. In the original work, \cite{schonsheck2018parallel} determines this vector field as the gradient of the geodesic distance function on the mesh $\mathcal{M}$. Thus, this requires the choice of a \textit{seed point} on the mesh with which to compute the distance to. Then the first axis of the local frame corresponds to the gradient of the geodesic, while the second axis corresponds to the cross product of this gradient and the surface normal. Motivated by \cite{boscaini2016learning} which defines frames via principal directions of curvature, in our work, we use the first principal lines of curvature to define our vector field. To do this, we use the geodesic vector field to fix the ambiguity in the first principal directions of curvature so that both vectors lie in the same half-plane. We note that this global ambiguity is not necessarily resolved in \cite{boscaini2016learning}.

\section{Framework for Disentanglement via Geometry}
\label{sec:framework}
As discussed in the paper, the Fundamental Theorem of Surfaces motivates the use of our conformal factor and normal (CFAN) feature. To this point, we will formally state this theorem also known as the Bonnet Theorem. From this theorem, we will see that CFAN is essentially approximating the solution to a difficult system of partial differential equations known as the Gauss-Codazzi equations to reconstruct encoded surfaces. To establish this, we first introduce important geometric objects known as the fundamental forms. We refer to \cite{do2016differential} as the standard reference in classical differential geometry. 

\subsection{Fundamental Forms}
One of the most important concepts in the differential geometry of surfaces is that of fundamental forms. The first and second fundamental forms define the metric and embedded properties of a surface. We will define and relate these fundamental forms to identity and pose. 

\subsubsection{First Fundamental Form}
The first fundamental form is critical for understanding metric properties of a surface. It is used to determine the arclength of a curve on a surface. Additionally, it is used to determine the local angle between curves on a surface. 

Consider the point $p$ on the surface $\mathcal{M}$. Assuming the surface is regular, there exists some local parameterization of the surface, $r(u, v)$. Then $\{r_u, r_v\}$ is a natural basis on the tangent plane to the surface at point $p$, $T_p(\mathcal{M})$. Now consider two tangent vectors, $u_1 = (a, b)$ and $u_2= (c, d)$, where coordinates are with respect to the local basis. Then the inner product of these vectors is defined with the first fundamental form, $I$, as 
\begin{align}
    I_p(u_1, u_2) &= \begin{bmatrix} a & b  \end{bmatrix} \begin{bmatrix} E & F \\ F & G \end{bmatrix} \begin{bmatrix} c \\ d \end{bmatrix} \\
    &\text{where} \quad E = r_u \cdot r_u, F = r_u \cdot r_v, G = r_v \cdot r_v.
    \notag
\end{align}

As in Euclidean space, the inner product is precisely what gives a concept of length, area, and angle. This matrix is also denoted as $g$ and referred to as the metric tensor. For a standard basis in Euclidean space, the metric tensor is simply the identity, $\mI$. 

Now we consider what it means for two surfaces to share the same identity. A reasonable expectation is that local angle and local surface area will be the same for both surfaces if they have the same identity. Under this expectation, these surfaces are isometric and share the same first fundamental form. Thus, we appeal to the first fundamental form when evaluating the identity of a mesh. 

It is important to stress that this expectation assumes that a surface cannot go under any elastic deformation without altering the identity as this changes the metric. So this framework is not appropriate for certain datasets such as meshes of human face expression, as a change in expression involves relatively large scale elastic deformations of the face.   

\subsubsection{Second Fundamental Form}
We start by motivating the second fundamental form. Let $z = r_3(u,v)$ be a surface where the $uv$-plane is tangent to the surface at the origin. Then the Taylor expansion of $z$ about the origin is
\begin{align}
\label{eq:second_fund}
z = \frac{1}{2} \begin{bmatrix} u & v \end{bmatrix}
\begin{bmatrix} L & M \\ M & N
    \end{bmatrix}
    \begin{bmatrix} u \\ v \end{bmatrix} + \text{ h.o.t.} \\
\text{where} \quad L = r_{uu} \cdot \vn, M = r_{uv} \cdot \vn, N = r_{vv} \cdot \vn \notag
\end{align}
with $\vn$ denoting the surface normal at the origin. Then, the matrix in the equation, $II_p$, is referred to as the second fundamental form. Intuitively, the second fundamental form represents the second order deviation of the surface from the tangent plane at a given point.

Clearly, the second fundamental form depends on the surface normal $\vn$. Due to its dependence on the embedding of the surface in the ambient space, $\mathbb{R}^{3}$, the second fundamental form is not an intrinsic geometric property of the surface, and is hence extrinsic. This is unlike the first fundamental form which is invariant under the ambient embedding of the surface. We associate this embedding into Euclidean space with the pose of the surface. 

\subsection{Gauss-Codazzi Equations}

Before we can introduce the fundamental theorem, we must first define the Gauss-Codazzi equations. The Gauss equations consist of 
\begin{align} \label{eq:gauss}
    EK &= (\Gamma_{11}^1)_2 - (\Gamma_{12}^2)_1 + \Gamma_{11}^1 \Gamma_{12}^2 + \Gamma_{11}^2 \Gamma_{22}^2  - \Gamma_{12}^1 \Gamma_{11}^2 - (\Gamma_{12}^2)^2  \\
    FK &= (\Gamma_{12}^1)_1 - (\Gamma_{11}^1)_2 + \Gamma_{12}^2 \Gamma_{12}^1  - \Gamma_{11}^2 \Gamma_{22}^1 \notag \\
    FK &= (\Gamma_{12}^2)_2 - (\Gamma_{22}^2)_1 + \Gamma_{12}^1 \Gamma_{12}^2  - \Gamma_{22}^1 \Gamma_{11}^2 \notag \\
    GK &= (\Gamma_{22}^1)_1 - (\Gamma_{12}^1)_2 + \Gamma_{22}^1 \Gamma_{11}^1 + \Gamma_{22}^2 \Gamma_{12}^1 - \Gamma_{12}^2 \Gamma_{22}^1 - (\Gamma_{12}^1)^2. \notag \\
    \text{where} &\quad K := \frac{LN - M^2}{EG - F^2} \notag
\end{align}

In terms of classical differential geometry, the Codazzi-Mainardi equations are defined as 
\begin{align} \label{eq:gc}
    L_v - M_u &= L \Gamma_{12}^1 + M( \Gamma_{12}^2 - \Gamma_{11}^1) - N \Gamma_{11}^2 \\
    M_v - N_u &= L \Gamma_{22}^1 + M( \Gamma_{22}^2 - \Gamma_{12}^1) - N \Gamma_{12}^2. \notag
\end{align}
Together, these form the Gauss-Codazzi equations. Here $\Gamma$ denotes the Christoffel symbols of the second kind. These symbols are defined using Einstein summation notation as 
\begin{equation}
    \Gamma_{kl}^i := \frac{1}{2} g^{im} \left(\frac{\partial g_{mk}}{\partial x^l} + \frac{\partial g_{ml}}{\partial x^k} - \frac{\partial g_{kl}}{\partial x^m}\right),
\end{equation}
where $g$ is the metric tensor, and $g^{im} := (g^{-1})_{im}$. The operator $\frac{\partial}{\partial x^i}$ refers to the partial derivative with respect to the $i^{th}$ coordinate of the local basis. Also, we note that the Christoffel symbols are symmetric in their lower indices. 

Since the metric tensor $g$ and the first fundamental form $I$ are synonymous, the Gauss-Codazzi equations is a system of partial differential equations in terms of the first and second fundamental forms. Now, we formally state Bonnet's classical result.

\begin{theorem}[Fundamental Theorem of Surfaces]
Let $U$ be a simply connected domain in $\mathbb{R}^2$ and let $E, F, G, L, M,$ and $N$ be functions in $C^{\infty}(U)$. Additionally, we assume $E$ and $F > 0$, and $EG - F^2 >0$. Lastly, these functions satisfy the Gauss-Codazzi equations (\ref{eq:gauss}, \ref{eq:gc}). Then there exists an immersion $f: U \rightarrow \mathbb{R}^3$ with first and second fundamental forms
\begin{align}
    I &= E du^2 + 2 F du dv + G dv^2 \\
    II &= L du^2 + 2 M du dv + N dv^2 \notag.
\end{align}
The immersion $f$ is unique up to rotation and translation.
\end{theorem}

In total, this theorem establishes that a surface can be reconstructed up to rigid motion given its first and second fundamental forms. Additionally, these fundamental forms satisfy a compatibility conditions known as the Gauss-Codazzi equations. As established in the main body of the paper, for genus zero surfaces, the conformal factor with respect to some reference surface defines the first fundamental form of a surface up to scaling. Additionally, it is clear from \eqref{eq:second_fund} that the second fundamental form can be recovered from the first fundamental form and the surface normals. Thus, it follows that a genus zero surface can be reconstructed given the CFAN feature, assuming compatibility conditions are satisfied.  

Though the fundamental theorem concludes that a surface can be reconstructed given its fundamental forms, it is hard to reconstruct this surface in practice. This is because the reconstruction depends on solving the Gauss-Codazzi equations which are clearly complex. To this end, CFAN-VAE can be viewed as approximating the solution to the Gauss-Codazzi equations, essentially solving this system of partial differential equations. With this, we gain a better understanding of the problem that CFAN-VAE is solving and a better appreciation for its performance.    

\section{Additional Details on Numerical Experiments}
\label{sec:add_setup}

\subsection{Concerning Datasets}
\label{subsec:app_datasets}
We provide additional details on the SMAL and SURREAL datasets used in this work. As mentioned, \cite{Zuffi:CVPR:2017} makes use of the SMPL model from \cite{Loper2015SMPLAS} to generate data. SMPL is a skinning model which allows for the generation of synthetic data given a shape parameter and a pose parameter. In the SMAL dataset, we have a mean shape parameter associated with five classes of animals; cats, dogs, horses, cows, and hippos. For each class, we randomly sample 4,000 pose vectors and shape vectors from a normal distribution with standard deviation of 0.2. Then we generate a mesh for each shape parameter, added to the mean shape parameter for the associated class, and each pose parameter. These generated meshes form our training, validation, and test sets with split $(15000, 2000, 3000)$.

The SURREAL dataset was formed in the method of \cite{groueix20183d}. Here we generate 20,000 male and female meshes via sampling a set of SURREAL parameters for training. We sample another 1,500 of each category for validation and testing. We repeat this with 3,000 and 300 meshes per training and non-training for bent meshes. Thus, our data split is $(46000, 600, 3000)$. As mentioned earlier, we add diffuse the SURREAL $\beta$ parameter so that the dataset contains similar but non-isometric meshes. 

\subsection{Concerning Architectures}
\label{subsec:app_architecture}

We provide details on the widths of our networks used for training. These widths are the number of convolution kernels per layer. They are as follows for the specified datasets and latent dimensions $(\vz_c, \vz_n)$:
\begin{itemize}
    \item DFAUST (32, 32); SURREAL (8, 8), (16, 16), (32, 32)
    \subitem CFAN-VAE Encoders (12, 24, 48, 48, 96)
    \subitem xyz-VAE Encoders (16, 32, 64, 64, 128)
    \subitem Decoder (128, 64, 64, 32, 16)
    \item SURREAL (5, 15)
    \subitem CFAN-VAE Encoders (24, 48, 96, 96, 192)
    \subitem xyz-VAE Encoders (32, 64, 128, 128, 256)
    \subitem Decoder (256, 128, 128, 64, 32)
    \item SMAL (8, 32)
    \subitem Conformal Encoder (24, 48, 96, 96, 192)
    \subitem Normal Encoder (48, 96, 96, 96, 192)
    \subitem xyz-VAE Encoder (32, 64, 128, 128, 256)
    \subitem CFAN-VAE Decoder (256, 128, 128, 128, 64)
    \subitem xyz-VAE Decoder (256, 128, 128, 64, 32)
\end{itemize}


\begin{figure*}[tb]
    \centering
    \begin{subfigure}[b]{0.29\textwidth}
      \includegraphics[width=\textwidth]{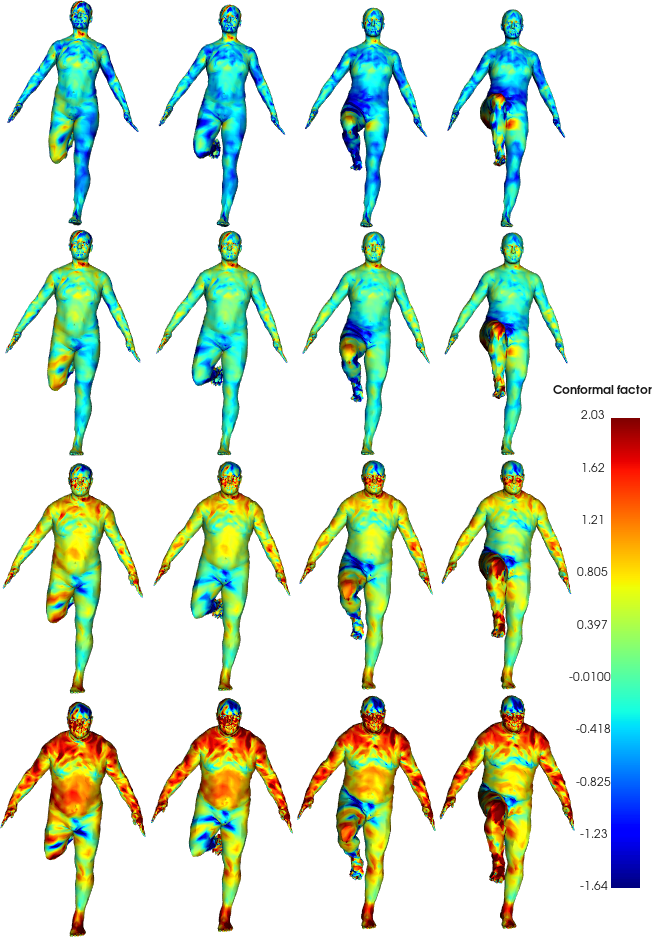}
      \end{subfigure}
    \hfill
    \begin{subfigure}[b]{0.19\textwidth}
      \includegraphics[width=\textwidth]{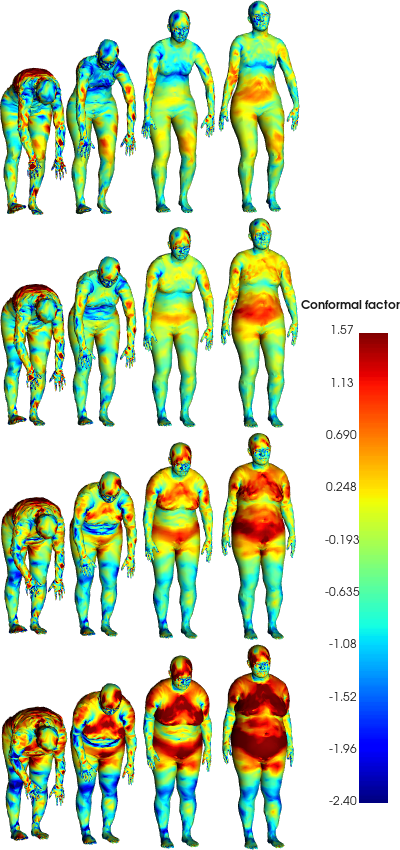}
      \end{subfigure}
    \hfill 
    \begin{subfigure}[b]{0.41\textwidth}
      \includegraphics[width=\textwidth]{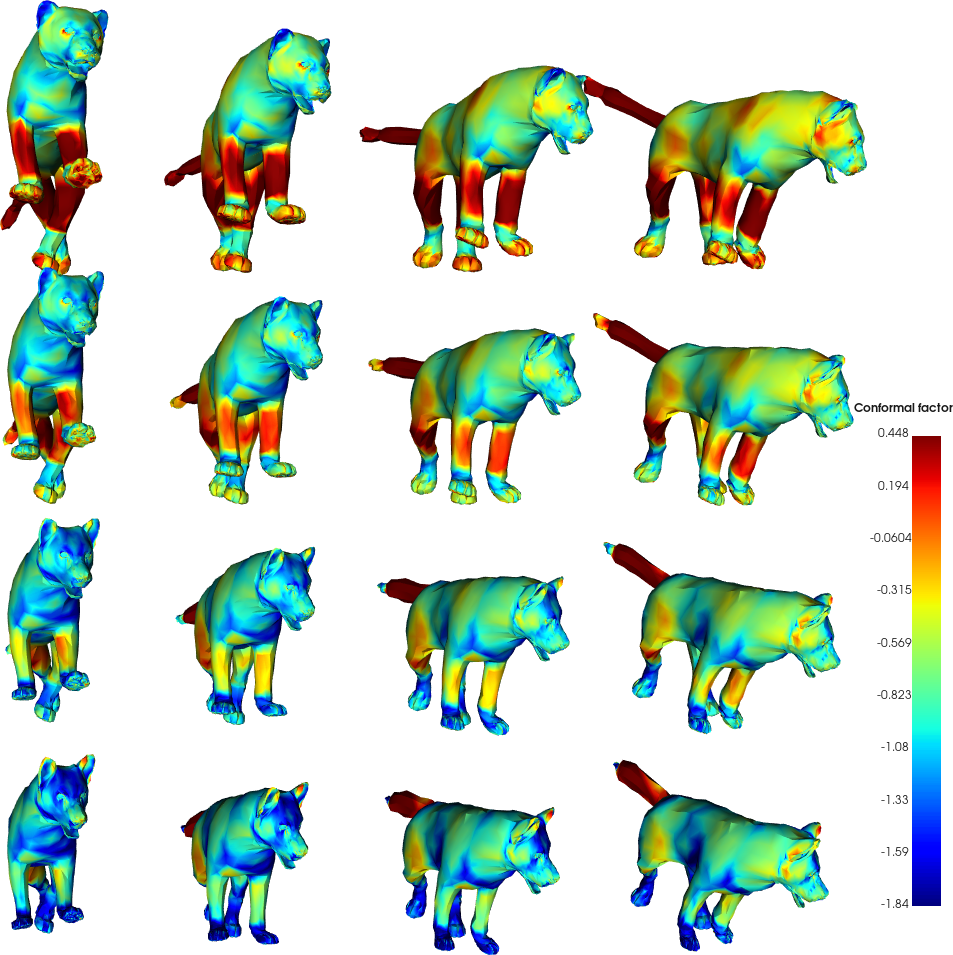}
      \end{subfigure}
    \caption{Additional geometrically disentangled interpolations using CFAN-VAE as in Figure \ref{fig:interp}.}
    \label{fig:add_interp_cfan}
\end{figure*}

\section{Additional Figures}
\label{sec:add_figures}

This section contains additional figures demonstrating the qualitative performance of CFAN-VAE. Specifically, we display additional instances of disentangled interpolation. Videos displaying interpolations can be found in the supplementary material folder for each dataset.

\begin{figure*}[tb]
    \centering
    \begin{subfigure}[b]{0.31\textwidth}
      \includegraphics[width=\textwidth]{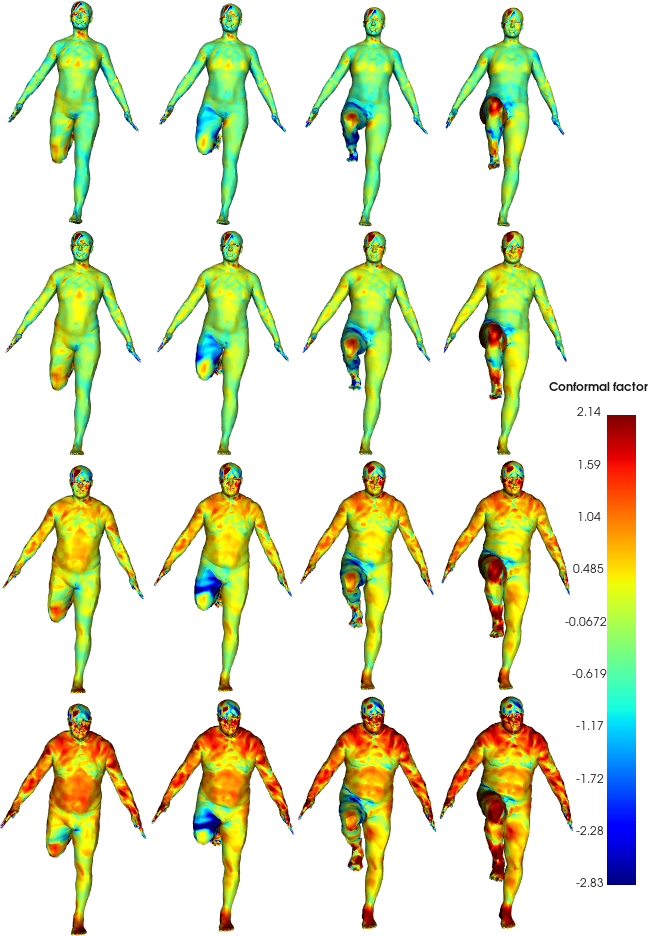}
      \end{subfigure}
    \hfill
    \begin{subfigure}[b]{0.21\textwidth}
      \includegraphics[width=\textwidth]{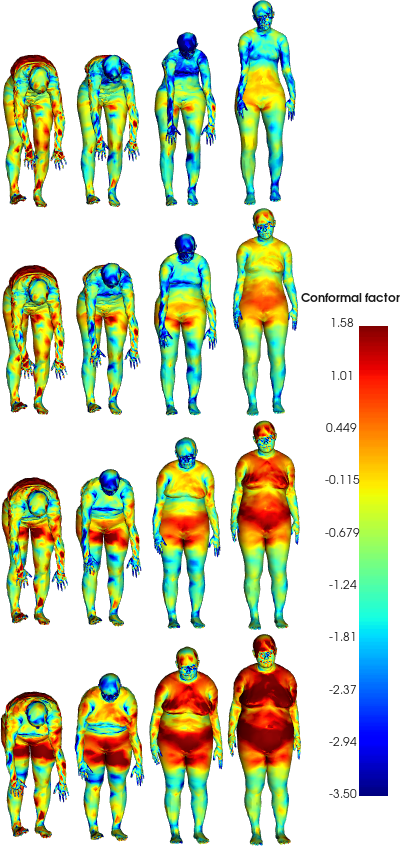}
      \end{subfigure}
    \hfill 
    \begin{subfigure}[b]{0.37\textwidth}
      \includegraphics[width=\textwidth]{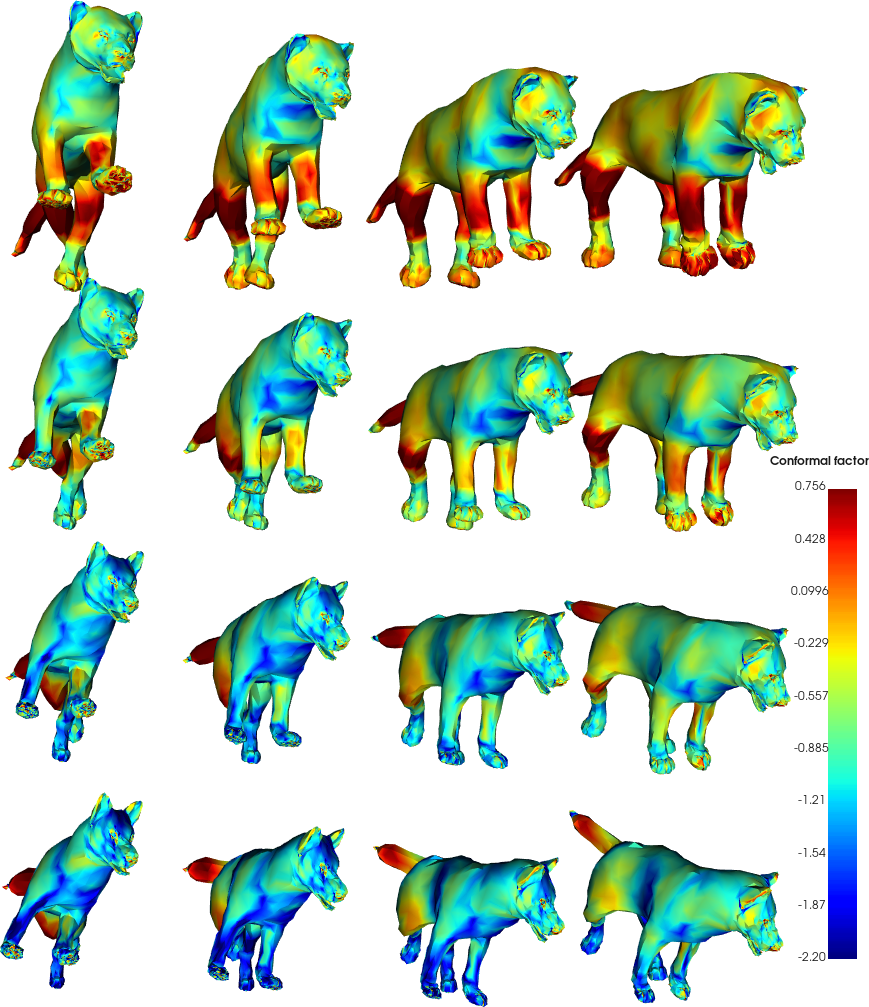}
      \end{subfigure}
    \caption{Additional geometrically disentangled interpolations using the registered xyz-VAE as in Figure \ref{fig:interp_xyz}.}
    \label{fig:add_interp_xyz}
\end{figure*}

\end{document}